\documentclass[10pt,twocolumn,letterpaper]{article}

\usepackage[pagenumbers]{cvpr} 

\usepackage[dvipsnames]{xcolor}

\usepackage{graphicx}
\usepackage{amsmath}
\usepackage{amssymb}
\usepackage{booktabs} 
\usepackage{xcolor,colortbl}
\usepackage{xspace}
\usepackage{makecell}

\usepackage{multirow}
\usepackage[linesnumbered,ruled,vlined]{algorithm2e}

\usepackage{pgfplots}
\DeclareUnicodeCharacter{2212}{−}
\usepgfplotslibrary{groupplots,dateplot}
\usetikzlibrary{patterns,shapes.arrows}
\pgfplotsset{compat=newest}
\pgfplotsset{every tick label/.append style={font=\footnotesize}}
\pgfplotsset{legend style={font=\footnotesize}}

\newcommand{\simplex}[1]{\Delta^{#1}}
\def\bo{{\boldsymbol{\omega}}}

\usepackage{xfrac}
\newcommand*\circled[1]{\tikz[baseline=(char.base)]{
            \node[shape=circle,draw,inner sep=.6pt] (char) {#1};}}

\makeatletter
\newcommand{\printfnsymbol}[1]{%
  \textsuperscript{\@fnsymbol{#1}}%
}
\makeatother

\definecolor{cvprblue}{rgb}{0.21,0.49,0.74}
\definecolor{method}{RGB}{222,187,195}
\usepackage[pagebackref,breaklinks,colorlinks,citecolor=cvprblue]{hyperref}

\title{Estimating calibration error under label shift without labels}

\author{Teodora Popordanoska\thanks{Equal contribution.}
\and 
Gorjan Radevski\printfnsymbol{1} 
\and
Tinne Tuytelaars 
\and
Matthew B. Blaschko 
\and \\
ESAT-PSI\\
KU Leuven, Belgium\\
{\tt\small \{firstname\}.\{lastname\}@kuleuven.be}
}

\begin{document}
\maketitle

\begin{abstract} 
In the face of dataset shift, model calibration plays a pivotal role in ensuring the reliability of machine learning systems.  
Calibration error (CE) is an indicator of the alignment between the predicted probabilities and the classifier accuracy.
While prior works have delved into the implications of dataset shift on calibration, existing CE estimators assume access to labels from the target domain, which are often unavailable in practice, i.e., when the model is deployed and used. 
This work addresses such challenging scenario, and proposes a novel CE estimator under \emph{label shift}, which is characterized by changes in the marginal label distribution $p(Y)$, while keeping the conditional $p(X|Y)$ constant between the source and target distributions.
Our contribution is an approach, which, by leveraging importance re-weighting of the labeled source distribution, provides consistent and asymptotically unbiased CE estimation with respect to the shifted target distribution.
Empirical results across diverse real-world datasets, under various conditions and label-shift intensities, demonstrate the effectiveness and reliability of the proposed estimator.
\end{abstract}

\section{Introduction}
\begin{figure}[t]
    \centering
    \resizebox{1.0\columnwidth}{!}{
    \includegraphics{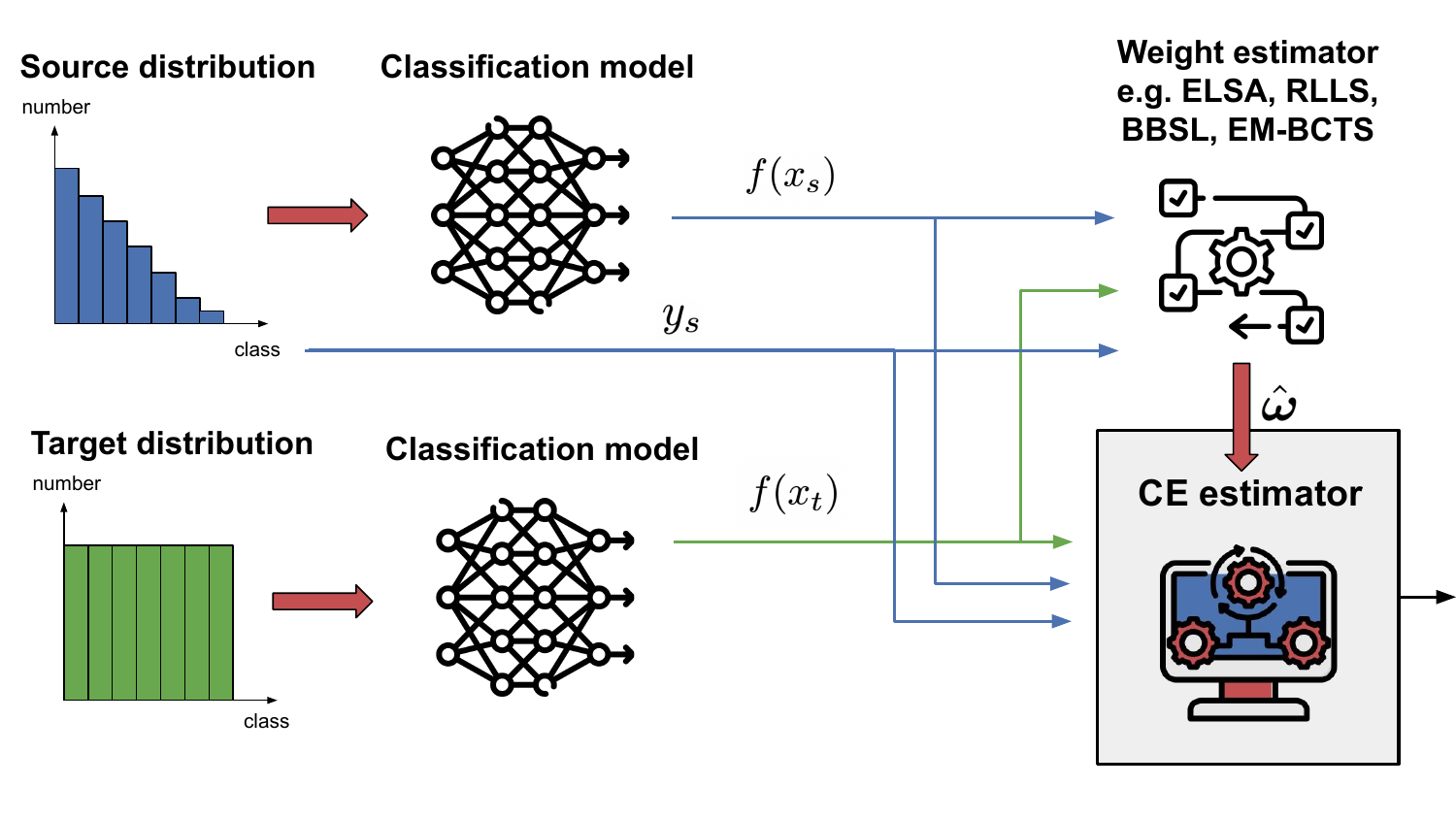}
    }
    \caption{Overview of the proposed method. We consider a label shift setting (e.g. the source follows a long-tail distribution, and the target is balanced). Our CE estimator leverages importance weights to account for the shift in the target distribution. }
    \label{fig:figure1}
\end{figure}

Reliable uncertainty estimation plays a pivotal role in predictive models, especially in safety-critical applications, where decisions based on neural network predictions can have significant consequences~\cite{kompa2021second}. The \textit{calibration error} (CE) \cite{naeini2015,guo2017calibration,vaicenavicius2019evaluating} of a model provides insights into the reliability and trustworthiness of its predictions. Informally, CE is a measure of discrepancy between predicted probabilities and empirically observed class frequencies. For instance, when a $calibrated$ model predicts an 80\% chance that a patient has the flu, we expect that out of 100 patients with similar symptoms, 80 indeed have the flu.

While numerous approaches exist for estimating CE and mitigating miscalibration, all of them rely on the availability of i.i.d.\ and labeled hold-out data. However, these assumptions are frequently violated in practical settings. 
On one hand, the source (train) distribution may differ from the target (test) distribution -- a phenomenon known as \textit{dataset shift}~\cite{quinonero2008dataset}.
On the other hand, the assumption of having access to labeled target data for estimating CE is often unrealistic or excessively costly. For domains like medical diagnostics during disease outbreaks, acquiring labeled patient data is essential for accurate calibration estimation, but may be prohibitively expensive due to the need for expert annotations and the urgency of the situation.

Motivated by this problem, where classifier trained on historical data must adapt to scenarios with varying class prevalences, we focus on one of the most common types of dataset shift -- \textit{label shift} \cite{moreno2012unifying,quinonero2008dataset}. Label shift refers to the scenario where the marginal distribution of the labels changes from source to target, i.e., $p_s(Y)\neq p_t(Y)$, while the conditional distribution of features given the label remains the same $p_s(X| Y)=p_t(X| Y)$. It corresponds to anti-causal learning (predicting the cause $y$ from its effects $x$ \cite{scholkopf2012causal}). For example, during a pneumonia outbreak, $p(Y)$ (e.g. flu) might rise, but the symptoms of the disease $p(X| Y)$ (e.g. cough given flu) do not change. 

In this work, we tackle the challenge of measuring calibration in the presence of \textit{label shift}, without access to labels on the target data. In particular, we propose \textit{the first consistent and asymptotically unbiased CE estimator} for such setting. We build on ideas from unsupervised domain adaptation, and employ importance weighting (via current state-of-the-art methods) to estimate the degree of shift in the target distribution.
Furthermore, we derive the variance of the CE estimator, both in the case when labels are available (e.g., when estimating CE on source data), and in the label-shifted scenario. An overview of our method is shown in Fig.~\ref{fig:figure1}. To the best of our knowledge, no other CE estimator exists for assessing calibration under label shift without labels.
We conduct thorough experiments across a variety of datasets, models, weight estimators, intensities of shift on the target distribution, and imbalance factors of the source distribution to validate the performance of our estimator. The experiments showcase the capability of our approach to effectively assess calibration in such context.

\textbf{Contributions:} \circled{1} We propose and derive the first CE estimator under label shift which \textit{does not require} labels from the label-shifted target distribution (\S\ref{sec:methods}); \circled{2} We demonstrate the effectiveness of the estimator on standard datasets (\S\ref{sec:effectiveness}), and show that it yields reliable CE estimates across different models and importance weights estimators; \circled{3} On real-world datasets (\S\ref{sec:real-data}), we ablate various aspects of the estimator, including the influence of the sample size and the impact of varying ratios of labeled source and unlabeled target samples. The results confirm the robustness of the method across many different scenarios.

\section{Related work}
\textbf{Estimating CE} is a challenging task in machine learning, since it requires
estimating an expectation conditioned on a continuous random variable, i.e., $\mathbb{E} \left[ Y \mid f \left( X \right) \right]$, where $Y$ is a one-hot encoded label, $X$ is an input variable and $f$ is a probabilistic model. 
The most common approach to estimate CE is based on binning \cite{zadrozny2001obtaining,naeini2015}.
In the \textit{binary} classification setting, the unit interval $[0, 1]$ is typically split into intervals (bins) of equal width \cite{nguyen2015posterior}. However, the predictions of a trained neural network are usually non-uniform, resulting in many bins with very few samples. Thus, an alternative approach was proposed \cite{vaicenavicius2019evaluating}, which partitions the probability simplex such that equal number of points fall in every bin. This is so-called equal mass (or adaptive) binning scheme. The number of bins and the binning scheme can significantly influence the estimated value \cite{kumar2019verified}, and there is no optimal default since every setting has a different bias-variance trade-off \cite{nixon2019measuring}. 
The prevalent measure to quantify calibration of a \textit{multi-class} model is known as expected calibration error (ECE) \cite{naeini2015,guo2017calibration}. Typically, it is used to assess the weakest form of calibration, top-label (or confidence) calibration \cite{guo2017calibration}, which only considers the confidence score of the predicted class. Class-wise calibration \cite{kull2019beyond} is a stronger notion, which requires calibrated scores for each class. This involves comparing $f_k(X)$ with $\mathbb{E} \left[ Y_k \mid f_k \left( X \right) \right]$ for each class $k$. Both notions depend on estimating a conditional distribution given a univariate continuous random variable, which allows the estimation schemes of one notion to be applied to the other. Canonical calibration \cite{vaicenavicius2019evaluating,popordanoska2022} is the strictest notion, as it requires the whole probability vector to be calibrated, i.e., $f(X)$ should match $\mathbb{E} \left[ Y \mid f \left( X \right) \right]$. 
In this work, we focus on binary and class-wise CE, estimated using an adaptive binning scheme.

\textbf{Label shift}, also known as prior probability shift~\cite{lipton2018detecting, azizzadenesheli2019regularized, alexandari2020maximum} is often intertwined with the broader concept of unsupervised domain adaptation \cite{kouw2021review}. Several different methods have been proposed to address label shift: importance re-weighting \cite{lipton2018detecting,azizzadenesheli2019regularized, saerens2002adjusting,tian2023elsa}, kernel mean matching (KMM)~\cite{zhang2013domain}, and methods based on generative adversarial training~\cite{guo2020label}. In general, there are two popular importance re-weighting approaches: one based on maximizing the likelihood function and the other based on inverting a confusion matrix. Using the first approach, Saerens \etal \cite{saerens2002adjusting} proposed an Expectation Maximization (EM) procedure to estimate the shift in the class priors between the source and target distributions. The advantage of EM is that it requires neither retraining, nor hyperparameter tuning. However, it assumes that the predictions are calibrated, which is often not the case for modern neural networks \cite{guo2017calibration}. To overcome this limitation, hybrid methods combining calibration techniques and domain adaptation methods have been proposed. For instance, Alexandari \etal\ \cite{alexandari2020maximum} propose Bias-Corrected Temperature Scaling (BCTS) 
alongside EM. 
Following the second approach, Lipton  \etal\ \cite{lipton2018detecting} proposed Black-Box Shift Learning (BBSL), which aims to estimate the re-weighting coefficients even in cases where the model is poorly calibrated. As an improvement over the BBSL method, Azizzadenesheli  \etal\ \cite{azizzadenesheli2019regularized} propose a technique with good statistical guarantees, known as Regularized Learning under Label Shifts (RLLS). This method introduces a regularization hyperparameter, designed to address the high estimation error of the importance weights in the low target sample regime. Both BBSL and RLLS estimate importance weights from a confusion matrix estimated on a held-out validation set. Furthermore, both methods can cope with label shift even when the classifier does not output calibrated probabilities, but they require model retraining with the importance weights. 
Recently, a moment-matching framework \cite{tian2023elsa} was proposed for addressing label shift called Efficient Label Shift Adaptation (ELSA). In the absence of target labels, our CE estimator relies on these importance weight estimators to re-weight the source data. 

\section{Methods}\label{sec:methods}
We consider a supervised classification setting and we use $X \in \mathcal{X} = \mathbb{R}^d$ and $Y \in \mathcal{Y} = \{1, \dots, k\}$ to denote the input and target variables. The data consists of two parts: labeled source data $\left\{(x_i, y_i) \right\}_{i=1}^n$ and unlabeled target data $\left\{ x_i \right\}_{i=n+1}^{n+m}$. Let $p_s$ and $p_t$ denote the source and target distributions, respectively. The support on the target label distribution is a subset of $\mathcal{Y}$, i.e., the target data does not contain new classes that are not in the source data.

Let us consider a probabalistic classifier $f \colon \mathcal{X} \to \simplex{k}$, where $\simplex{k} := \left\{ \left(p_1, \dots, p_k \right)^\intercal \in \left[0, 1 \right]^k \mid \sum_{i=1}^k p_i = 1 \right\}$ is a ($k-1$)-dimensional probability simplex over $k$ classes.  
Considering a case where $\mathcal{Y} =\{0, 1\}$, the \emph{binary calibration error} \cite{brocker_2009reliability,zadrozny2002,kumar2019verified} of a classifier $f$ with respect to a given $L_p$ space is given by:
\begin{equation}
    \label{eq:binary_calibration}
    \operatorname{BCE}_p \left( f \right) = \mathbb{E} \left[ \left( \mathbb{E} \left(Y = 1 \mid f \left( X \right) \right) - f \left( X \right) \right)^p \right] ^\frac{1}{p}.
\end{equation}
A straightforward extension to the multi-class setting is to define a \emph{class-wise calibration error} \cite{kull2019beyond,kumar2019verified,nixon2019measuring} as:
\begin{align}
    \operatorname{CWCE}_p \left( f \right) = \left( \frac{1}{k} \sum_{i=1}^k \mathbb{E} \left[ \left( \mathbb{E} \left(Y = i \mid f_i \left( X \right) \right) - f_i \left( X \right) \right)^p \right] \right)^\frac{1}{p}.
\end{align}
The strongest notion of calibration -- canonical -- can be assessed via an \textit{$L_p$ canonical calibration error} given by:
\begin{equation}
    \label{eq:canonical_calibration_error}
     \operatorname{CE}_p(f) = \biggl(\mathbb{E}\biggl[\Bigl\| \mathbb{E}[Y \mid f(X)]-f(X)\Bigr\|_p^p\biggr] \biggr)^{\frac{1}{p}}.
\end{equation}
Each of these calibration errors are defined with respect to a data distribution. It is therefore not surprising that the calibration of the model decreases under domain shift, which has also been empirically shown by several previous works \cite{ovadia2019can,karandikar2021soft}. However, related works estimate the calibration error on the shifted data using labels from the target domain, which are often not available in practice. Therefore, in the following section, we develop an estimator for the $L_p$ canonical CE under one of the most common types of shift -- \textit{label shift}. It is straigthforward to extend the canonical estimator to the class-wise and binary calibration error.

\subsection{Calibration error estimator}

The empirical estimator of Equation~\eqref{eq:canonical_calibration_error} w.r.t. the source distribution (no shift) is given by \cite{popordanoska2022}:
\begin{equation}
    \label{eq:general_canonical_estimator_no_shift}
   \widehat{\operatorname{CE}_{p}(f)^p} =  \frac{1}{n} \sum_{j=1}^{n} \left\| \widehat{\mathbb{E}_{p_s}[Y \mid f(x_j)]}- f(x_j) \right\|_p^p.
\end{equation}

Under dataset shift, assuming we have access to $n$ labeled samples from the source distribution and $m$ unlabeled samples from the target distribution, we wish to find an estimator of the form:
\begin{equation}
    \label{eq:general_canonical_estimator}
   \widehat{\operatorname{CE}_{p}(f)^p} =  \frac{1}{m} \sum_{j=n+1}^{m+n} \left\| \widehat{\mathbb{E}_{p_t}[Y \mid f(x_j)]}- f(x_j) \right\|_p^p, 
\end{equation}
where the expectations are taken w.r.t. the target data.

We consider a label shift setting where $p_s(Y) \neq p_t(Y)$ and $p_s(X \mid Y) = p_t(X \mid Y)$. A standard assumption in this case is the absolute continuity of the target label distribution with respect to the source distribution, i.e., for every $y \in \mathcal{Y}$ with $p_t(Y) > 0$, we require $p_s(Y) > 0$ \cite{lipton2018detecting}. 
In other words, the support of the target label distribution should be contained within the support of the source label distribution. 

The main challenge is to estimate the conditional expectation $\widehat{\mathbb{E}_{p_t}[Y \mid f(x_j)]}$ without having access to labels from the target distribution. Our approach makes use of importance weights $\bo=(\omega_1,\dots,\omega_k)$, where $\omega_i := p_t(Y=i)/p_s(Y=i)$, which would be used to re-weight the source label distribution.
We can estimate $p_s(Y=i)$. However, due to the lack of target labels, we cannot simply take a ratio of the empirical estimates. Thus, we rely on methods from the unsupervised domain adaptation literature \cite{tian2023elsa, alexandari2020maximum, azizzadenesheli2019regularized,lipton2018detecting}, to estimate the importance weights.
Then, for the conditional expectation, we have:
\begin{align}
   \mathbb{E}_{p_t}[Y \mid f(X)] &= \sum_{y} y \frac{p_t(Y=y, f(X) = f(x))}{p_t(f(X) = f(x))} \notag \\
   &= \sum_{y} y \frac{p_t(f(X)|Y)p_t(Y)}{p_t(f(X) = f(x))} \notag \\
   &= \sum_{y} y \frac{p_s(f(X)|Y)p_t(Y)}{p_t(f(X) = f(x))} \notag \\
   &= \sum_{y} y \frac{p_s(f(X)|Y)p_s(Y) \bo }{p_t(f(X) = f(x))} \notag \\
   &\approx \frac{\frac{1}{n} \hat{\bo} \sum_{i=1}^n k(f(X), f(x_i))y_i}{\frac{1}{m} \sum_{i=n+1}^{m+n} k(f(X), f(x_i))}
\end{align}
where $\bo=\frac{p_t(Y)}{p_s(Y)}$, $\hat{\bo}$ is its empirical estimate, and $k$ is any consistent kernel over its domain \cite{silverman1986density}. Keep in mind that we can estimate $p_t(f(X)|Y)$ with $p_s(f(X)|Y)$ because of the label shift assumption that the conditional expectation $p(X|Y) $ remains the same, and $f$ is a trained model (constant). The weights $\hat{\bo}$ are estimated for every $Y \in \mathcal{Y}$ with labeled source data, unlabeled target data and a trained predictor $f$. 
Plugging this back into Equation~\eqref{eq:general_canonical_estimator}, for CE under label shift we get:

\begin{equation}
    \label{eq:canonical_estimator_prior_shift}
   \widehat{\operatorname{CE}_{p}(f)^p} =  \frac{1}{m} \sum_{j=n+1}^{m+n} \left\| \hat{R_j}  - f(x_j) \right\|_p^p, 
\end{equation}

where 
\begin{equation}
    \hat{R_j} = \frac{\frac{1}{n} \hat{\bo} \sum_{i=1}^n k(f(x_j), f(x_i))y_i}{\frac{1}{m - 1} \sum_{\substack{i=n+1 \\ i \neq j}}^{m+n} k(f(x_j), f(x_i))}.
\end{equation}

The estimator has values $\in [0, 2]$, it is consistent and asymptotically unbiased \cite{popordanoska2022}. Depending on the choice of kernel, it can also be made differentiable and integrated as part of post-hoc and trainable calibration methods. For instance, \cite{popordanoska2022} and \cite{zhang2020mix} proposed Dirichlet and Triweight kernels, respectively. Here, in order to facilitate variance estimates in the next section, we propose using a binning kernel defined as:
\begin{equation}
    \label{eq:binning_kernel}
    k(f(x_i), f(x_j)) = \begin{cases}
        1 & \text{if $f(x_i)$ and $f(x_j)$} \\
          & \text{fall in the same bin} \\
        0 & \text{otherwise}
    \end{cases} .
\end{equation}
In the next part, we will derive the variance of the CE estimator, both w.r.t the source data (no shift), given in Equation~\eqref{eq:general_canonical_estimator_no_shift}, and w.r.t. the label shifted target distribution, with the estimator defined in Equation~\eqref{eq:canonical_estimator_prior_shift}.

\subsection{Variance of the estimator}
For simplicity, let us consider a binary classification setting with no label shift. The extension to the class-wise case is straightforward as we can apply the variance estimator per class, whereas the derivation for the label shift scenario is analogous and we defer it to  Appendix~\ref{appendix:variance}. Assuming an adaptive binning scheme with $b$ bins (same number of points fall in each bin), with a binning kernel $k$ as defined in Equation~\eqref{eq:binning_kernel}, we may simplify the empirical ratio in Equation~\eqref{eq:general_canonical_estimator_no_shift}:
\begin{equation}
    \hat{R}_j = \frac{b}{n-1} \sum_{i \in B_{f(x_j)}} y_i, 
\end{equation}
where $B_{f(x_j)}$ denotes the bin into which $f(x_j)$ is assigned.
Further, considering that Y is a Bernoulli random variable, and each bin contains $\frac{n-1}{b}$ points, we have that:
\begin{align}
    \sum_{i \in B_{f(x_j)}} y_i \sim &\operatorname{Binom}(R_j,\frac{n-1}{b}) \\
    \approx& \mathcal{N}\left(\frac{n-1}{b} R_j,\frac{n-1}{b} R_j(1-R_j)\right).
    \notag
\end{align}
This in turn implies for large $n$ that:
\begin{equation}
    \label{distribution_R}
    \hat{R}_j \sim \mathcal{N}\left( R_j,\frac{b}{n-1} R_j(1-R_j)\right) 
\end{equation}

Then, denoting $Z_j = |R_j - x_j|^p $, we may compute:
\begin{align}
    \label{eq:variance_no_shift}
    \operatorname{Var}\left(\frac{1}{n} \sum_{j=1}^n Z_j \right) &=
    \frac{1}{n^2} \operatorname{Var} \left(\sum_{j=1}^n Z_j \right) \\
    &= \frac{1}{n^2} \left( \sum_{j=1}^n \operatorname{Var} \left( Z_j \right) + \sum_{j \neq i} \operatorname{Cov} \left( Z_j, Z_i \right) \right) \notag
\end{align}
The procedure for computing Equation~\eqref{eq:variance_no_shift} is outlined in Algorithm~\ref{alg:calculate_variance}. With this development, we have for the first time an estimator of the variance of the CE estimator.

\begin{algorithm}
\caption{Calculate Variance}
\label{alg:calculate_variance}
\SetAlgoLined
\KwData{Confidence scores $f(x)$, labels $y$, $L_p$ norm $p$}
\KwResult{Variance of $\widehat{\operatorname{CE}_{p}(f)^p}$}

Initialize $inner\_sum$ to $0$\;
Initialize $bin\_samples$ to $\frac{n-1}{b}$

\For{$bin \leftarrow 1$ \KwTo $n\_bins$}{
    $f(x)_{bin}$, $y_{bin}$ $\leftarrow$ data falling into the current bin\;

    \For{$j \leftarrow 1$ \KwTo number of samples in the bin}{
        Sample $f(x_j)$ from $f(x)_{bin}$\;
        Sample $R_j$ per Equation~\eqref{distribution_R}\;
        Append $\left| R_j - x_j \right|^p$ to $samples\_for\_variance$\;
    }
    Calculate empirical $variance$ based on  $samples\_for\_variance$\;
    Calculate empirical $covariance$ based on data pairs within the bin\;
    Update $inner\_sum$ using $variance$ and $covariance$\;
}
Calculate final $variance$ as $\frac{inner\_sum}{n^2}$ \;
\Return{$variance$}\;
\end{algorithm}

\section{Experiments \& Discussion}
We broadly organize our experiments as follows. First, in \S\ref{sec:effectiveness}, on standard natural image datasets \cite{krizhevsky2009learning, deng2009imagenet}, using models of increasing complexity (i.e., depth), we verify the overall effectiveness of the proposed CE estimator under label shift imposed in different scenarios. Further, we ablate how our method works with various importance weights estimators \cite{tian2023elsa,alexandari2020maximum,azizzadenesheli2019regularized,lipton2018detecting}. In \S\ref{sec:real-data}, we conduct in-depth empirical analysis of various aspects of our method on three realistic scenarios featuring different modalities: tumor identification across hospital image images, text sentiment classification across different users, and animal wild trap species recognition from images. Across all experiments, we report results with our method using estimated weights in \textbf{\textcolor{method}{red}}. 
The code will be released upon acceptance.

\subsection{Effectiveness of the estimator}
\label{sec:effectiveness}
\textbf{Datasets.} We use CIFAR-10/100 Long Tail (LT) datasets \cite{cao2019learning}, which are simulated from CIFAR  \cite{krizhevsky2009learning} with an imbalance factor (IF) that controls the degree of shift. We also use the ImageNet-LT dataset \cite{liu2019largescale} -- simulated from ImageNet \cite{deng2009imagenet} -- allowing us to verify how well our method performs in a setting with a large number of diverse classes. To ensure that the source train and validation set are the same, we merge the long-tail train and the balanced validation set of ImageNet-LT, and then divide it into new train and validation datasets using stratified sampling, as per Chen \etal \cite{chen2023transfer}. We keep the balanced target test set unchanged. 

\textbf{Models.} We use ResNet \cite{he2016deep} models intialized from scratch with different depth (20, 34, 56, 110, 152). Please refer to Appendix~\ref{app:experiments} for details about the training process.

\textbf{Metrics.} We measure accuracy and $L_2$ CE \cite{kumar2019verified}. We use the subscripts $s$ and $t$ to differentiate whether the metric is calculated w.r.t.\ the source or target distribution, respectively. We denote the estimated CE using our method as $\widehat{\operatorname{CE}}$, the ground truth weights (assuming labels on both source and target) by $\hat{\bo}^*$, and the estimated weights by $\hat{\bo}$. Across all experiments, we fix the number of bins to 15 (in \S\ref{sec:real-data} we ablate the influence of this hyperparameter), and ensure that all bins have equal number of samples (i.e.\ adaptive binning \cite{vaicenavicius2019evaluating}). In multi-class settings, we measure class-wise CE and report the average across classes. Across all experiments, we report CE $\times 100$.

\textbf{Experimental setup.} 
We examine two different types of label shift in our experiments. In the first, the source dataset (train and validation set) follows a long-tail distribution among classes, while the target follows a uniform distribution (i.e., is balanced). In the second, the source is balanced, while the target exhibits a long-tail.
To impose the long-tail, we consider several imbalance factors, defined as a ratio of the number of samples in the largest and in the smallest class. 
Unless otherwise stated, we use the RLLS weight estimator to compute $\hat{\bo}$. We refer to Appendix~\ref{app:experiments} for implementation details.

\paragraph{Measuring calibration error under label shift.} We train ResNet models on the CIFAR-10/100 long-tail variants, and use an imbalance factor of 10 -- the least frequent class is subsampled to 10\% of the original size. We report performance in Table~\ref{table:cifarlt-imagenetlt}. On ImageNet-LT we resample the source classes so that each has frequency between 20 and 50 samples\footnote{We want to test on the classes belonging to the long-tail of frequencies. However, we empirically find that less than 20 samples per class in the source distribution leads to unreliable importance weight estimates.}.
\begin{table*}[t]
\centering
\resizebox{0.8\textwidth}{!}{
\begin{tabular}{ccccccc>{\columncolor{method}}c} \toprule
   {Dataset} & {Model} & {Accuracy$_s$} &  {Accuracy$_t$} & {$\operatorname{CE}_s$} & {$\operatorname{CE}_t$} & {$\widehat{\operatorname{CE}}_t(\hat{\bo}^*)$ } & \textbf{$\widehat{\operatorname{CE}}_t (\hat{\bo})$} \\ \midrule
    \multirow{4}{*}{CIFAR-10-LT}
     & ResNet-20  & 83.10 & 78.24  & 0.81 & 0.88  & 0.91  & 0.89 \\
     & ResNet-32  & 85.48 & 80.74  & 0.90 & 1.04  & 1.16  & 1.21 \\
    & ResNet-56  & 85.38 & 81.71  & 0.98 & 1.12  & 1.16  & 1.16 \\
    & ResNet-110  & 84.94 & 81.29  & 1.00 & 1.19  & 1.22  & 1.22 \\ 
    \midrule
    \multirow{4}{*}{CIFAR-100-LT} 
      & ResNet-20  & 52.24 & 44.81  & 0.61 & 0.66  & 0.66  & 0.64 \\
    & ResNet-32  & 53.48 & 47.73  & 0.66 & 0.71  & 0.71  & 0.70 \\
    & ResNet-56  & 54.21 & 47.18  & 0.66 & 0.72  & 0.72 & 0.71 \\
    & ResNet-110  & 56.58 & 49.78  & 0.69 & 0.73  & 0.73 & 0.72 \\ 
    \midrule
    \multirow{3}{*}{ImageNet-LT} 
    & Resnet18 & 33.56 & 27.65 & 0.034 & 0.037 & 0.037 & 0.037 \\
    & Resnet50 & 39.86 & 33.29 & 0.050 & 0.056 & 0.056 & 0.056 \\
    & Resnet152 & 43.20 & 36.90 & 0.052 & 0.059 & 0.059 & 0.059 \\
    \bottomrule
\end{tabular}
}
\caption{Performance evaluation of various models trained on long-tail distributed datasets. The target data follows a uniform distribution over the classes, while the source is obtained with a IF of 0.1. Our estimator $\widehat{\operatorname{CE}}_t (\hat{\bo})$ consistently provides reliable estimates of $\operatorname{CE}_t$.}
\label{table:cifarlt-imagenetlt}
\end{table*}
As expected, we observe that the models' accuracy significantly drops across different models and datasets, while the calibration error increases on the label shifted target distribution. We also observe that our method yields reliable calibration error estimates, which is consistent across model depths and datasets. Importantly, the estimate with our method improves when the weights are obtained as ground truth ratios between the source and target data. This suggests a room for improvement of our estimator as weight estimation methods continue to improve.

Most existing works on importance weights estimation \cite{tian2023elsa,alexandari2020maximum,lipton2018detecting, azizzadenesheli2019regularized} predominantly focus on an alternative type of label shift, where the source is balanced (i.e., the classes have equal frequency), while the target follows a long-tail distribution. We report results for such settings in Table~\ref{table:balanced-cifar}, where we induce label shift on the target data with imbalance factors of increasing magnitude: 1 (balanced), 1.25, 2, 10, 100. As before, we perform experiments with ResNet models of varying depths to verify that our findings generalize across models of different complexities. Our results on both CIFAR-10/100 reveal that our method yields reliable calibration error estimates in the absence of labeled target data, irrespective of the IF intensity.

\begin{figure}
    \centering
    \resizebox{0.95\columnwidth}{!}{
\begin{tikzpicture}

\definecolor{darkslategray38}{RGB}{38,38,38}
\definecolor{darkslategray66}{RGB}{66,66,66}
\definecolor{indianred1819295}{RGB}{181,92,95}
\definecolor{lavender234234242}{RGB}{234,234,242}
\definecolor{mediumseagreen95157109}{RGB}{95,157,109}
\definecolor{peru20313699}{RGB}{203,136,99}
\definecolor{steelblue88116163}{RGB}{88,116,163}

\begin{axis}[
axis background/.style={fill=lavender234234242},
axis line style={white},
scaled x ticks = false,
scaled y ticks = false,
tick align=outside,
tick pos=left,
unbounded coords=jump,
x grid style={white},
x tick label style={/pgf/number format/fixed},
xmajorticks=false,
xmajorticks=true,
xmin=-0.5, xmax=3.5,
xtick style={color=darkslategray38},
xtick={0,1,2,3},
xticklabels={ELSA,BBSL,RLLS,EM-BCTS},
y grid style={white},
y tick label style={/pgf/number format/.cd, fixed, fixed zerofill, precision=3},
ylabel=\textcolor{darkslategray38}{$ \left|\operatorname{CE}_t - \widehat{\operatorname{CE}_t}\right|$},
ymajorgrids,
ymajorticks=false,
ymajorticks=true,
ymin=0, ymax=0.10395,
ytick style={color=darkslategray38}
]
\draw[draw=black,fill=steelblue88116163] (axis cs:-0.3,0) rectangle (axis cs:0.3,0.0427999999999999);
\draw[draw=black,fill=peru20313699] (axis cs:0.7,0) rectangle (axis cs:1.3,0.099);
\draw[draw=black,fill=mediumseagreen95157109] (axis cs:1.7,0) rectangle (axis cs:2.3,0.0103);
\draw[draw=black,fill=indianred1819295] (axis cs:2.7,0) rectangle (axis cs:3.3,0.0161);

\end{axis}

\end{tikzpicture}
    }
    \caption{Comparison of several common weight estimators using ResNet-110 on CIFAR-100-LT. The RLLS method performs the best when incorporated in our CE estimator, as measured by the absolute difference between estimated and ground truth CE. }
    \label{fig:weight_estimators}
\end{figure}
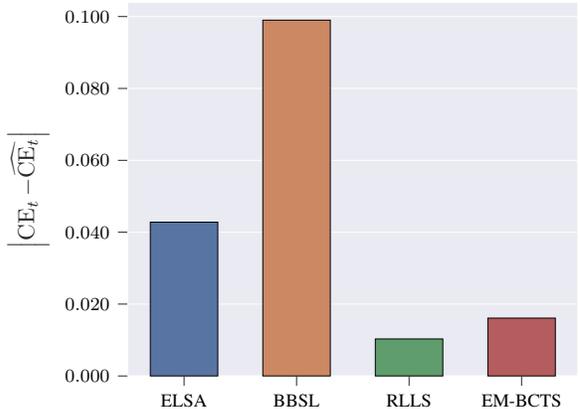

\textbf{Influence of the weight estimation method.} Note that our proposed CE estimator relies on the availability of per-class importance weights. Such weights can be obtained using domain adaptation methods under label shift. In Fig.~\ref{fig:weight_estimators} we compare the performance of several common weight estimators (ELSA, RLLS, BBSL, EM-BCTS) when incorporated in our CE estimator. We showcase results using ResNet-110 on CIFAR-100-LT, however, we observed similar trends across multiple settings. We observe that RLLS emerges as most favorable compared to the others, providing reasonable importance weight estimates, which results in overall optimal performance of our estimator. We henceforth report all results using estimated weights with RLLS. See Appendix~\ref{app:experiments} for detailed experiments involving other weight estimators.

\begin{table*}[t]
\centering
\resizebox{0.8\textwidth}{!}{
\begin{tabular}{ccccc>{\columncolor{method}}cc>{\columncolor{method}}cc>{\columncolor{method}}cc>{\columncolor{method}}c} \toprule
    \multicolumn{4}{l}{\textbf{\textit{Imbalance factor (IF) intensity}}} & \multicolumn{2}{c}{\textbf{IF=1.25}} & \multicolumn{2}{c}{\textbf{IF=2.0}} & \multicolumn{2}{c}{\textbf{IF=10.0}} & \multicolumn{2}{c}{\textbf{IF=100.0}} \\
   {Dataset} & {Model} & {$\operatorname{CE}_s$} & {$\operatorname{CE}_t$ } & { $\operatorname{CE}_t$} & {$\widehat{\operatorname{CE}_t}$} & { $\operatorname{CE}_t$} & {$\widehat{\operatorname{CE}_t}$} & { $\operatorname{CE}_t$} & {$\widehat{\operatorname{CE}_t}$} & { $\operatorname{CE}_t$} & {$\widehat{\operatorname{CE}_t}$} \\ 
   \midrule
    \multirow{4}{*}{CIFAR-10}
    & \text{ResNet-20} & 0.88 & 0.88 & 0.84 & 0.85 & 0.54 & 0.54 & 0.82 & 0.84 & 1.09 & 1.15 \\
    & \text{ResNet-32} & 1.08 & 1.07 & 1.03 & 1.03 & 0.65 & 0.66 & 0.90 & 0.94 & 1.05 & 1.05 \\
    & \text{ResNet-56} & 1.00 & 1.01 & 0.95 & 0.95 & 0.60 & 0.61 & 0.86 & 0.88 & 1.07 & 1.08 \\
    & \text{ResNet-110} & 1.39 & 1.38 & 1.27 & 1.28 & 0.77 & 0.79 & 1.03 & 1.04 & 1.05 & 1.06 \\
    \midrule
    \multirow{4}{*}{CIFAR-100} 
    & \text{ResNet-20} & 0.59 & 0.59 & 0.59 & 0.60 & 0.59 & 0.60 & 0.58 & 0.58 & 0.54 & 0.53 \\
    & \text{ResNet-32} & 0.67 & 0.67 & 0.67 & 0.68 & 0.67 & 0.67 & 0.65 & 0.66 & 0.62 & 0.61 \\
    & \text{ResNet-56} & 0.74 & 0.75 & 0.75 & 0.75 & 0.75 & 0.75 & 0.74 & 0.74 & 0.69 & 0.70 \\
    & \text{ResNet-110} & 0.76 & 0.76 & 0.76 & 0.76 & 0.76 & 0.76 & 0.74 & 0.74 & 0.68 & 0.68 \\ 
    \bottomrule
\end{tabular}
}
\caption{Performance evaluation of our estimator on a label-shifted target distribution using different imbalance factors. The models are trained on a balanced CIFAR-10/100. Across all shifts, our estimator yields accurate estimates compared to the ground truth (with labels), and effectively handles even the most severe case with $\operatorname{IF}=100$.}
\label{table:balanced-cifar}
\end{table*}

\subsection{Label shifted CE estimation in the wild}\label{sec:real-data}
\textbf{Datasets.} We use real world datasets from WILDS \cite{koh2021wilds} with different input modalities: Camelyon17 \cite{bandi2018detection} and iWildCam \cite{beery2021iwildcam} -- images, and Amazon \cite{ni2019justifying} -- text. Camelyon17 consists of histopatological images of a patient lymph node section with potential metastatic breast cancer. The labels are binary and denote whether the central region contains a tumor. The iWildCam dataset consists of images from animal traps in the wild, while the labels are different animal species. The Amazon dataset contains review text inputs paired with 1-out-of-5 star ratings as labels. 

\textbf{Models.} For experiments on iWildCam we report results with a standard ResNet-50 \cite{he2016deep}, two ViT transformer-based models \cite{dosovitskiy2020image} (large variant with standard image resolution -- 224, and large variant with increased image resolution -- 384),  and Swin-Large \cite{liu2021swin} (all pre-trained on ImageNet). For experiments on the Amazon dataset, we use pre-trained transformer-based models -- BERT \cite{devlin2018bert}, RoBERTa \cite{liu2019roberta}, Distill-Bert \cite{sanh2019distilbert} (D-BERT) and Distill-Roberta (D-ROBERTA)\footnote{Obtained from RoBERTa with the same procedure as Sahn \etal \cite{sanh2019distilbert}.}. For the ablation studies on Camelyon17, we use a ResNet-50 pre-trained on ImageNet.

\textbf{Experimental setup.} The iWildCam, Amazon and Camelyon17 datasets have an i.i.d.\ validation set, serving as our source distribution. Additionally, iWildCam and Amazon also have i.i.d.\ test set, to which we apply label shift and use as our target distribution. On iWildCam, we select the 20 most-frequent classes from the test dataset, and based on the frequency of the least frequent class, we randomly subsample each class to obtain a uniform target distribution. On Amazon, we subsample the test data based on the least frequent class to obtain label-shifted target data which is uniformly distributed among classes.
Due to the absence of such a test set for Camelyon17, we partitioned the original train set to form a target set, whose label distribution is shifted with various intensity levels in the experiments.

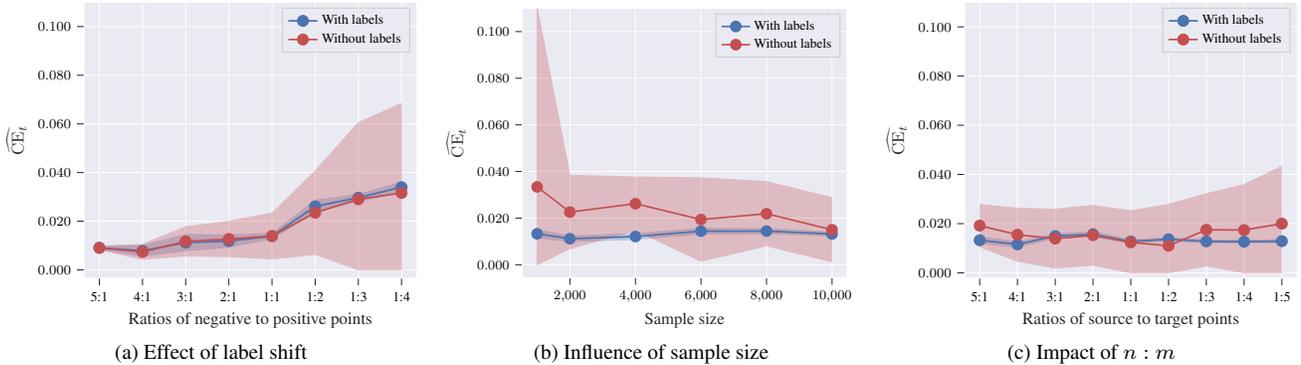
\begin{figure*}[t]
    \centering
    \subfloat[Effect of label shift]{
    \resizebox{0.32\textwidth}{!}{
\begin{tikzpicture}

\definecolor{darkslategray38}{RGB}{38,38,38}
\definecolor{indianred1967981}{RGB}{196,79,81}
\definecolor{lavender234234242}{RGB}{234,234,242}
\definecolor{lightgray204}{RGB}{204,204,204}
\definecolor{steelblue76114176}{RGB}{76,114,176}

\begin{axis}[
axis background/.style={fill=lavender234234242},
axis line style={white},
legend cell align={left},
legend style={
  fill opacity=1,
  draw opacity=1,
  text opacity=1,
  draw=lightgray204,
  fill=lavender234234242
},
scaled x ticks = false,
scaled y ticks = false,
tick align=outside,
tick pos=left,
x grid style={white},
x tick label style={/pgf/number format/fixed},
xlabel=\textcolor{darkslategray38}{Ratios of negative to positive points},
xmajorgrids,
xmajorticks=false,
xmajorticks=true,
xmin=-0.35, xmax=7.35,
xtick style={color=darkslategray38},
xtick={0,1,2,3,4,5,6,7},
xticklabels={5:1,4:1,3:1,2:1,1:1,1:2,1:3,1:4},
y grid style={white},
y tick label style={/pgf/number format/.cd, fixed, fixed zerofill, precision=3},
ylabel=\textcolor{darkslategray38}{$\widehat{\operatorname{CE}_t}$},
ymajorgrids,
ymajorticks=false,
ymajorticks=true,
ymin=-0.0034214546673514, ymax=0.11,
ytick style={color=darkslategray38}
]
\path [draw=steelblue76114176, fill=steelblue76114176, opacity=0.3, line width=0pt]
(axis cs:0,0.0097668596371083)
--(axis cs:0,0.00845660139828363)
--(axis cs:1,0.00559074260317813)
--(axis cs:2,0.00778995836882943)
--(axis cs:3,0.00929812481312946)
--(axis cs:4,0.0126721171135581)
--(axis cs:5,0.0235623408196604)
--(axis cs:6,0.0283105363576882)
--(axis cs:7,0.0318427094387723)
--(axis cs:7,0.0361298461849155)
--(axis cs:7,0.0361298461849155)
--(axis cs:6,0.0310225281639314)
--(axis cs:5,0.0286274267954775)
--(axis cs:4,0.015039355286001)
--(axis cs:3,0.0143950787244297)
--(axis cs:2,0.0145778117337566)
--(axis cs:1,0.0100500383563196)
--(axis cs:0,0.0097668596371083)
--cycle;

\path [draw=indianred1967981, fill=indianred1967981, opacity=0.3, line width=0pt]
(axis cs:0,0.00971258921326859)
--(axis cs:0,0.00828741078673141)
--(axis cs:1,0.00431941000497706)
--(axis cs:2,0.00568910494733316)
--(axis cs:3,0.0054371228257196)
--(axis cs:4,0.00447723658261546)
--(axis cs:5,0.00632880772555499)
--(axis cs:6,0)
--(axis cs:7,0)
--(axis cs:7,0.068429093347028)
--(axis cs:7,0.068429093347028)
--(axis cs:6,0.0605203703462701)
--(axis cs:5,0.040671192274445)
--(axis cs:4,0.0233227634173845)
--(axis cs:3,0.0199628771742804)
--(axis cs:2,0.0177108950526668)
--(axis cs:1,0.0104805899950229)
--(axis cs:0,0.00971258921326859)
--cycle;

\addplot [line width=1pt, steelblue76114176, mark=*, mark size=3, mark options={solid}]
table {%
0 0.00911173051769596
1 0.00782039047974887
2 0.011183885051293
3 0.0118466017687796
4 0.0138557361997796
5 0.0260948838075689
6 0.0296665322608098
7 0.0339862778118439
};
\addlegendentry{With labels}
\addplot [line width=1pt, indianred1967981, mark=*, mark size=3, mark options={solid}]
table {%
0 0.009
1 0.0074
2 0.0117
3 0.0127
4 0.0139
5 0.0235
6 0.029
7 0.0316
};
\addlegendentry{Without labels}
\end{axis}

\end{tikzpicture}}
    \label{subfig:ratios_vs_ece_formula_p2}
    }
    \subfloat[Influence of sample size]{
    \resizebox{0.32\textwidth}{!}{
\begin{tikzpicture}

\definecolor{darkslategray38}{RGB}{38,38,38}
\definecolor{indianred1967981}{RGB}{196,79,81}
\definecolor{lavender234234242}{RGB}{234,234,242}
\definecolor{lightgray204}{RGB}{204,204,204}
\definecolor{steelblue76114176}{RGB}{76,114,176}

\begin{axis}[
axis background/.style={fill=lavender234234242},
axis line style={white},
legend cell align={left},
legend style={
  fill opacity=1,
  draw opacity=1,
  text opacity=1,
  draw=lightgray204,
  fill=lavender234234242
},
scaled x ticks = false,
scaled y ticks = false,
tick align=outside,
tick pos=left,
x grid style={white},
x tick label style={/pgf/number format/fixed},
xlabel=\textcolor{darkslategray38}{Sample size},
xmajorgrids,
xmajorticks=false,
xmajorticks=true,
xmin=550, xmax=10450,
xtick style={color=darkslategray38},
y grid style={white},
y tick label style={/pgf/number format/.cd, fixed, fixed zerofill, precision=3},
ylabel=\textcolor{darkslategray38}{$\widehat{\operatorname{CE}_t}$},
ymajorgrids,
ymajorticks=false,
ymajorticks=true,
ymin=-0.0055168572200787, ymax=0.11,
ytick style={color=darkslategray38}
]
\path [draw=steelblue76114176, fill=steelblue76114176, opacity=0.3, line width=0pt]
(axis cs:1000,0.0149977627004489)
--(axis cs:1000,0.0115754547133864)
--(axis cs:2000,0.00994865591665253)
--(axis cs:4000,0.0108581648556476)
--(axis cs:6000,0.0132002163191592)
--(axis cs:8000,0.0133539987469516)
--(axis cs:10000,0.0123319279184471)
--(axis cs:10000,0.0140482337906611)
--(axis cs:10000,0.0140482337906611)
--(axis cs:8000,0.0155006812982639)
--(axis cs:6000,0.0156264341231716)
--(axis cs:4000,0.0133656859251428)
--(axis cs:2000,0.01236064755281)
--(axis cs:1000,0.0149977627004489)
--cycle;

\path [draw=indianred1967981, fill=indianred1967981, opacity=0.3, line width=0pt]
(axis cs:1000,0.110337144401574)
--(axis cs:1000,0)
--(axis cs:2000,0.0068423058947096)
--(axis cs:4000,0.0147562518138469)
--(axis cs:6000,0.00150488902897354)
--(axis cs:8000,0.00812620256600663)
--(axis cs:10000,0.00124698107913115)
--(axis cs:10000,0.0287530189208689)
--(axis cs:10000,0.0287530189208689)
--(axis cs:8000,0.0356737974339934)
--(axis cs:6000,0.0372951109710265)
--(axis cs:4000,0.0376437481861531)
--(axis cs:2000,0.0383576941052904)
--(axis cs:1000,0.110337144401574)
--cycle;

\addplot [line width=1pt, steelblue76114176, mark=*, mark size=3, mark options={solid}]
table {%
1000 0.0132866087069176
2000 0.0111546517347313
4000 0.0121119253903952
6000 0.0144133252211654
8000 0.0144273400226077
10000 0.0131900808545541
};
\addlegendentry{With labels}
\addplot [line width=1pt, indianred1967981, mark=*, mark size=3, mark options={solid}]
table {%
1000 0.0334
2000 0.0226
4000 0.0262
6000 0.0194
8000 0.0219
10000 0.015
};
\addlegendentry{Without labels}
\end{axis}

\end{tikzpicture}}
    \label{subfig:sample_size_vs_ece_formula_p2}
    }
    \subfloat[Impact of $n: m$]{
    \resizebox{0.32\textwidth}{!}{
\begin{tikzpicture}

\definecolor{darkslategray38}{RGB}{38,38,38}
\definecolor{indianred1967981}{RGB}{196,79,81}
\definecolor{lavender234234242}{RGB}{234,234,242}
\definecolor{lightgray204}{RGB}{204,204,204}
\definecolor{steelblue76114176}{RGB}{76,114,176}

\begin{axis}[
axis background/.style={fill=lavender234234242},
axis line style={white},
legend cell align={left},
legend style={
  fill opacity=1,
  draw opacity=1,
  text opacity=1,
  draw=lightgray204,
  fill=lavender234234242
},
scaled x ticks = false,
scaled y ticks = false,
tick align=outside,
tick pos=left,
x grid style={white},
x tick label style={/pgf/number format/fixed},
xlabel=\textcolor{darkslategray38}{Ratios of source to target points},
xmajorgrids,
xmajorticks=false,
xmajorticks=true,
xmin=-0.4, xmax=8.4,
xtick style={color=darkslategray38},
xtick={0,1,2,3,4,5,6,7,8},
xticklabels={5:1,4:1,3:1,2:1,1:1,1:2,1:3,1:4,1:5},
y grid style={white},
y tick label style={/pgf/number format/.cd, fixed, fixed zerofill, precision=3},
ylabel=\textcolor{darkslategray38}{$\widehat{\operatorname{CE}_t}$},
ymajorgrids,
ymajorticks=false,
ymajorticks=true,
ymin=-0.00216533871663345, ymax=0.11,
ytick style={color=darkslategray38}
]
\path [draw=steelblue76114176, fill=steelblue76114176, opacity=0.3, line width=0pt]
(axis cs:0,0.0146125815513233)
--(axis cs:0,0.0119190719274183)
--(axis cs:1,0.0104661037870298)
--(axis cs:2,0.0140654259875597)
--(axis cs:3,0.0146736030736461)
--(axis cs:4,0.0119941203731614)
--(axis cs:5,0.0129834922824049)
--(axis cs:6,0.0121981179668556)
--(axis cs:7,0.0120741245265539)
--(axis cs:8,0.0122459776509871)
--(axis cs:8,0.0135160904567371)
--(axis cs:8,0.0135160904567371)
--(axis cs:7,0.0132710500691597)
--(axis cs:6,0.0133929593161692)
--(axis cs:5,0.0143252399887896)
--(axis cs:4,0.0134934702080173)
--(axis cs:3,0.0167395534655556)
--(axis cs:2,0.0158344878300844)
--(axis cs:1,0.0126749152592768)
--(axis cs:0,0.0146125815513233)
--cycle;

\path [draw=indianred1967981, fill=indianred1967981, opacity=0.3, line width=0pt]
(axis cs:0,0.0278078535751106)
--(axis cs:0,0.0105921464248894)
--(axis cs:1,0.00471595365978845)
--(axis cs:2,0.00192682711996543)
--(axis cs:3,0.00314650390049496)
--(axis cs:4,0)
--(axis cs:5,0)
--(axis cs:6,0.00282259752514388)
--(axis cs:7,0)
--(axis cs:8,0)
--(axis cs:8,0.043306774332669)
--(axis cs:8,0.043306774332669)
--(axis cs:7,0.0358816802003705)
--(axis cs:6,0.0321774024748561)
--(axis cs:5,0.0278712964799176)
--(axis cs:4,0.0252186021681719)
--(axis cs:3,0.027453496099505)
--(axis cs:2,0.0258731728800346)
--(axis cs:1,0.0262840463402116)
--(axis cs:0,0.0278078535751106)
--cycle;

\addplot [line width=1pt, steelblue76114176, mark=*, mark size=3, mark options={solid}]
table {%
0 0.0132658267393708
1 0.0115705095231533
2 0.0149499569088221
3 0.0157065782696009
4 0.0127437952905893
5 0.0136543661355972
6 0.0127955386415124
7 0.0126725872978568
8 0.0128810340538621
};
\addlegendentry{With labels}
\addplot [line width=1pt, indianred1967981, mark=*, mark size=3, mark options={solid}]
table {%
0 0.0192
1 0.0155
2 0.0139
3 0.0153
4 0.0124
5 0.011
6 0.0175
7 0.0174
8 0.02
};
\addlegendentry{Without labels}
\end{axis}

\end{tikzpicture}}
    \label{subfig:ratios_nm_vs_ece_formula_p2}
    }
    \caption{The performance of our estimator under various conditions. The shaded region represents the standard deviation obtained from our variance formula and the steps outlined in Algorithm~\ref{alg:calculate_variance}. The estimator is able to generalize well to a wide range of label shifts, sample sizes, and ratios of source to target samples.}
    \label{fig:camelyon_experiments_p2}
\end{figure*}

\textbf{Results.}
We report results in Table~\ref{table:amazon-iwildcam} to estimate how our CE estimator performs in scenarios where label shift occurs on real-world data. Similarly to our previous observations, there is a significant drop in accuracy between the source and target distributions, as well as a considerable increase in CE when testing the models on label-shifted data. Importantly, we observe that for the problem of animal species recognition (iWildCam) and sentiment classification (Amazon) our CE estimates ($\widehat{\operatorname{CE}_t} (\hat{\bo})$) closely follow the estimates obtained using access to target labels (${\operatorname{CE}_t}$). Similarly, employing ground truth importance weights ($\widehat{\operatorname{CE}_t}(\hat{\bo}^*)$) improves the performance further.

\begin{table}[t]
\centering
\resizebox{1.0\columnwidth}{!}{
\begin{tabular}{ccccc>{\columncolor{method}}c} \toprule
   {Dataset} & {Model} & {${\operatorname{CE}_s}$} & {${\operatorname{CE}_t}$} & {$\widehat{\operatorname{CE}_t}(\hat{\bo}^*)$} & {$\widehat{\operatorname{CE}_t} (\hat{\bo})$} \\ 
   \midrule
    \multirow{4}{*}{Amazon} 
    & RoBERTa & 0.92 & 2.27 & 2.19 & 2.02 \\
    & D-RoBERTa & 1.71 & 3.50 & 3.41 & 3.36 \\
    & BERT &  0.29 & 1.25 & 1.30 & 1.06 \\ 
    & D-BERT &  2.37 & 4.37 & 4.65 & 4.41 \\ \midrule
    \multirow{3}{*}{iWildCam} 
    & ResNet-50 & 0.66 & 0.99 & 1.09 & 1.13 \\
    & ViT-Large & 0.75 & 1.08 & 1.26 & 1.38 \\
    & ViT-Large (384) & 0.77 & 1.14 & 1.23 & 1.27 \\
    & Swin-Large & 0.73 & 1.07 & 1.09 & 1.17 \\
    \bottomrule
\end{tabular}
}
\caption{Animal species image recognition and review-sentiment text classification on iWildCam and Amazon respectively.}
\label{table:amazon-iwildcam}
\end{table}

\subsection{Ablations: Label-shifted tumor identification}
We conduct a series of ablation studies using the Camelyon17 dataset. Our reasons for choosing this dataset are threefold: (i) the application is both realistic and safety-critical; (ii) the dataset is balanced across source and target, allowing us to alter both distributions as per the hypothesis we are trying to verify; (iii) the problem is binary, allowing us to study the estimator properties on a simple problem.

Across the experiments, we create the source train and validation sets by retaining all negative samples and subsampling a portion of the positives. Unless stated otherwise, we report results obtained by sampling $20\%$ of the positive samples for training, i.e., $5:1$ ratio of negative to positive points. In Fig.~\ref{fig:camelyon_experiments_p2} we compare the estimated CE values (with no labels) to the ground truth (with labels) across different experimental scenarios. The shaded region illustrates the standard deviation obtained using the variance (as derived in Algorithm~\ref{alg:calculate_variance}). 

\textbf{Effect of increasing the target distribution label shift.} In Fig.~\ref{subfig:ratios_vs_ece_formula_p2}, we impose a constraint such that the size of the source and target distribution is the same ($n = m$), and we systematically shift the target. We do so by modifying the ratio of negative to positive samples as: $5:1, 5:4, ..., 1:1, ..., 1:4$. Therefore, in the most favorable scenario ($5:1$), the source and target distribution are the same (no label shift), while in the extreme $1:4$, we have $4$ times as many positive samples in the target data (which could occur, e.g. during a disease outbreak). We first observe that the estimated CE closely follows the ground truth, even in the most extreme case. Furthermore, we observe that the variance increases with the intensity of the shift -- indicating greater uncertainty and reducing the confidence one should have in the CE estimates.

\textbf{Effect of the source and target data size.} In Fig.~\ref{subfig:sample_size_vs_ece_formula_p2}, we also constrain that $n = m$, and we incrementally vary the sample size from $1,000$ to $10,000$ samples. In practice, low data regimes are common where annotated data is costly to obtain. We observe that, expectedly, the CE estimates deviate from the ground truth the most when using the fewest source and target samples (i.e., $1000$), and improve as the quantity of labeled data increases.

\textbf{Effect of varying ratios of labeled source to unlabeled target data.} In Fig.~\ref{subfig:ratios_nm_vs_ece_formula_p2}, we verify whether changing the ratio of source to target samples ($n:m$) has an effect of the calibration error estimate. We observe that across different $n:m$ rations, our estimator yields reliable calibration error estimates, even in the most extreme scenario with $5 \times$ more target than source samples.

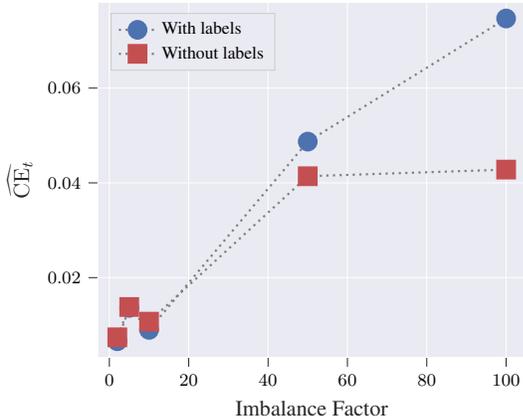
\begin{figure}
    \centering
    \resizebox{0.85\columnwidth}{!}{
\begin{tikzpicture}

\definecolor{darkslategray38}{RGB}{38,38,38}
\definecolor{gray}{RGB}{128,128,128}
\definecolor{indianred1967981}{RGB}{196,79,81}
\definecolor{lavender234234242}{RGB}{234,234,242}
\definecolor{lightgray204}{RGB}{204,204,204}
\definecolor{steelblue76114176}{RGB}{76,114,176}

\begin{axis}[
axis background/.style={fill=lavender234234242},
axis line style={white},
legend cell align={left},
legend style={
  fill opacity=1,
  draw opacity=1,
  text opacity=1,
  at={(0.03,0.97)},
  anchor=north west,
  draw=lightgray204,
  fill=lavender234234242
},
scaled x ticks = false,
scaled y ticks = false,
tick align=outside,
tick pos=left,
x grid style={white},
x tick label style={/pgf/number format/fixed},
xlabel=\textcolor{darkslategray38}{Imbalance Factor},
xmajorgrids,
xmajorticks=false,
xmajorticks=true,
xmin=-2.9, xmax=104.9,
xtick style={color=darkslategray38},
y grid style={white},
y tick label style={/pgf/number format/.cd, fixed},
ylabel=\textcolor{darkslategray38}{$\widehat{\operatorname{CE}_t}$},
ymajorgrids,
ymajorticks=false,
ymajorticks=true,
ymin=0.00316690571044992, ymax=0.0781042819241334,
ytick style={color=darkslategray38}
]
\addplot [line width=1pt, dotted, gray, mark=*, mark size=4, mark options={solid,fill=steelblue76114176,draw=steelblue76114176}]
table {%
100 0.0746980375507841
50 0.0487062430506057
10 0.00897151034207077
5 0.0136216344646426
2 0.00657315008379917
};
\addlegendentry{With labels}
\addplot [line width=1pt, dotted, gray, mark=square*, mark size=4, mark options={solid,fill=indianred1967981, draw=indianred1967981}]
table {%
100 0.0428
50 0.0414
10 0.0107
5 0.0138
2 0.0074
};
\addlegendentry{Without labels}
\end{axis}

\end{tikzpicture}}
    \caption{The effect of the imbalance factor of the source distribution on the quality of our estimator. }
    \label{fig:imb_factors}
\end{figure}

\textbf{Effect of the source imbalance factor.} In Fig.~\ref{fig:imb_factors}, we investigate the effect of the imbalance factor of the source distribution on calibration error estimate obtained by our estimator. To that end, we train a model on a long-tail source data with varying ratios of negative to positive samples, and we measure CE on a uniform target distribution. This is the opposite scenario from Fig.~\ref{subfig:ratios_vs_ece_formula_p2}, where we kept the source distribution fixed at $5:1$ and we varied the ratios of the target distribution. 
We observe that the CE deteriorates with the increase of IF. Similarly, our CE estimates increasingly deviate from the ground truth as the IF is increased. However, notice that in the most extreme case we explore ($\operatorname{IF}=100$), we train the model using $16,000$ negative samples and only $160$ positive samples.

\begin{figure}
    \centering
    \resizebox{0.85\columnwidth}{!}{
\begin{tikzpicture}

\definecolor{darkslategray38}{RGB}{38,38,38}
\definecolor{indianred1967981}{RGB}{196,79,81}
\definecolor{lavender234234242}{RGB}{234,234,242}
\definecolor{lightgray204}{RGB}{204,204,204}
\definecolor{steelblue76114176}{RGB}{76,114,176}

\begin{axis}[
axis background/.style={fill=lavender234234242},
axis line style={white},
legend cell align={center},
legend style={
  fill opacity=0.8,
  draw opacity=1,
  text opacity=1,
  draw=lightgray204,
  fill=lavender234234242
},
scaled x ticks = false,
scaled y ticks = false,
tick align=outside,
tick pos=left,
x grid style={white},
x tick label style={/pgf/number format/fixed},
xlabel=\textcolor{darkslategray38}{Number of Bins},
xmajorgrids,
xmajorticks=false,
xmajorticks=true,
xmin=-0.425, xmax=3.425,
xtick style={color=darkslategray38},
xtick={0,1,2,3},
xticklabels={10,15,20,50},
y grid style={white},
y tick label style={/pgf/number format/.cd, fixed, fixed zerofill, precision=3},
ylabel=\textcolor{darkslategray38}{$\widehat{\operatorname{CE}_t}$},
ymajorgrids,
ymajorticks=false,
ymajorticks=true,
ymin=0, ymax=0.025,
ytick style={color=darkslategray38}
]
\draw[draw=black,fill=steelblue76114176] (axis cs:0,0) rectangle (axis cs:0.25,0.0214990100630177);
\addlegendimage{ybar,ybar legend,draw=black,fill=steelblue76114176}
\addlegendentry{With labels}

\draw[draw=black,fill=steelblue76114176] (axis cs:1,0) rectangle (axis cs:1.25,0.0135370947717065);
\draw[draw=black,fill=steelblue76114176] (axis cs:2,0) rectangle (axis cs:2.25,0.0166080274632112);
\draw[draw=black,fill=steelblue76114176] (axis cs:3,0) rectangle (axis cs:3.25,0.0153894270252744);
\draw[draw=black,fill=indianred1967981] (axis cs:-0.25,0) rectangle (axis cs:0,0.0225);
\addlegendimage{ybar,ybar legend,draw=black,fill=indianred1967981}
\addlegendentry{Without labels}

\draw[draw=black,fill=indianred1967981] (axis cs:0.75,0) rectangle (axis cs:1,0.0146);
\draw[draw=black,fill=indianred1967981] (axis cs:1.75,0) rectangle (axis cs:2,0.016);
\draw[draw=black,fill=indianred1967981] (axis cs:2.75,0) rectangle (axis cs:3,0.0192);
\end{axis}

\end{tikzpicture}}
    \caption{The impact of the number of bins. The source distribution is imbalanced with $5:1$ ratio, whereas the target is uniform. Each binning setting has a different bias-vairance tradeoff. }
    \label{fig:n_bins}
\end{figure}
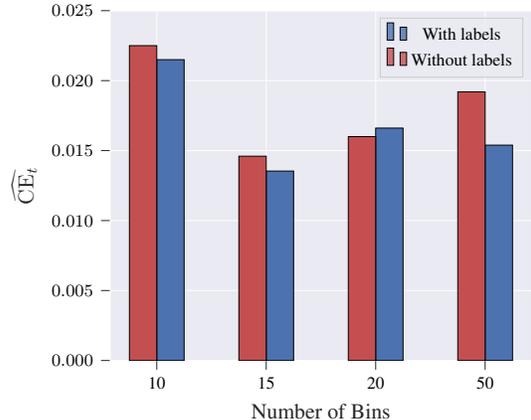

\textbf{Effect of the number of bins.} In Fig.~\ref{fig:n_bins} we see that for common choices for the number of bins, i.e., 10, 15 and 20 \cite{guo2017calibration}, the CE estimates closely align with the CE obtained using ground truth labels. We observe a higher discrepancy between the calibration error estimate and the ground truth value when using 50 bins.

In summary, the results offer a comprehensive understanding of the estimator's behavior and its ability to handle varying conditions. We can conclude that the estimator is able to generalize well to a wide range of settings and reliably estimate the calibration error of the target distribution, without requiring labels.

\section{Conclusion}
In this work, we addressed the problem of estimating CE of an \textit{unlabeled} target distribution under \textit{label shift}. We establish the first CE estimator and successfully measure CE in this challenging setting. Our proposed method accounts for the change in the label distribution by using importance-re-weighting of the source labels. 
As real-world data does not typically fulfill the label shift assumptions \cite{azizzadenesheli2019regularized}, we also derive the variance of this estimator, which provides critical insights into the reliability and stability of the estimates. We lay out a procedure for calculating the variance both when labels are available (e.g.\ on source), and in the label shift scenario when target labels are unavailable.

The proposed estimator has favorable statistical properties as it is consistent and asymptotically unbiased for any consistent kernel over its domain. We focus on a binning kernel, as this is the most commonly used form of the estimator for calibration error.
It is worth noting that choosing a differentiable kernel \cite{popordanoska2022} would make the estimator suitable for integrating into both post-hoc and trainable calibration methods, which we leave for future work.

We perform rigorous experimental analysis to evaluate the performance on different datasets, model architectures, intensities of label shift, weight estimators etc. Our experimental findings contribute to a nuanced understanding of the estimator's strengths and weaknesses. 

\textbf{Limitations.} Firstly, our estimator is specifically designed to address label-shift, however, other types of dataset-shift are equally important. Specifically, \emph{covariate shift} \cite{shimodaira2000_improving} is a common type of shift, where the marginal distribution over features changes, while the conditional distribution remains the same, i.e. $p_s(x) \neq p_t(x)$ and  $p_s(y \mid x) = p_t(y \mid x)$. However, each type of shift has its own intricacies and challenges and is beyond the scope of this paper.
We consider addressing this type of data-shift a crucial direction for future work. 
Further, our method depends on the accuracy of the importance weights, which we obtain from current state-of-the-art methods \cite{tian2023elsa,azizzadenesheli2019regularized,alexandari2020maximum}. Naturally, we inherit the limitations of such methods -- if, certain classes in the data are under-represented, the importance weights could be unreliable. For that reason, we also experiment with ground truth weights, showcasing that as importance weight estimation methods improve, our proposed estimator follows the same trend.

We consider this paper an important step toward assessing CE under shifts of this nature. By accounting for label shift, we can conduct more comprehensive and realistic evaluation of the model's calibration performance, which is especially important in safety-critical applications.

\section*{Acknowledgements}
This research received funding from the Flemish Government (AI Research Program) and the Research Foundation - Flanders (FWO) through project number G0G2921N.

{
    \small
    \bibliographystyle{ieeenat_fullname}
    \bibliography{main}
}

\appendix
\onecolumn

\begin{center}
    \Large\textbf{Supplementary Material}
\end{center}
   


\section{Variance of the CE estimator under label shift}
\label{appendix:variance}
\subsection{Theory}
In this section we derive the variance for the CE estimator under label shift, given in Equation~\eqref{eq:canonical_estimator_prior_shift}. We again consider a binary classification setting and assume an adaptive binning scheme with $b$ bins, such that an equal number of points fall into each bin. Since the numerator and the denominator in Equation~\eqref{eq:canonical_estimator_prior_shift} are computed w.r.t. different distributions (source and target), we may compute the bins boundaries according to either of them. For simplicity in the derivation, we assume that the bins are determined according to the target distribution. Then the empirical ratio simplifies to:
\begin{equation}
    \hat{R}_j = \frac{\frac{1}{n} \hat{\omega} \sum_{i \in Bin_{f(x_j)}} y_i}{\frac{1}{m-1} \frac{m-1}{b}} = \frac{b}{n} \hat{\omega} \sum_{i \in Bin_{f(x_j)}} y_j.
\end{equation}
Let $n_j$ denote the number of source points falling in the same bin as $f(x_j)$, i.e., $Bin_{f(x_j)}$. Then, using the Normal approximation to the Binomial distribution we obtain:
\begin{align}
    \sum_{i \in Bin_{f(x_j)}} y_i \sim &\operatorname{Binom}(p, n_j) \approx \mathcal{N}\left(n_j p, n_j p(1-p)\right) 
\end{align}
where $ p = \frac{1}{n_j} \sum_{i \in Bin_{f(x_j)}} y_i$. Subsequently, for large $n$, we may approximate the distribution of $\hat{R}_j$ with:
\begin{equation}
    \label{eq:ratio_label_shift}
    \hat{R}_i \sim \mathcal{N}\left(\frac{b}{n} \hat{\omega} n_j p, \frac{b^2}{n^2} \hat{\omega}^2 n_j p(1-p)\right).
\end{equation}

Finally, to compute the variance of the CE estimator under label shift, we replace Equation~\eqref{distribution_R} with Equation~\eqref{eq:ratio_label_shift} in Algorithm~\ref{alg:calculate_variance}, and the rest of the procedure remains the same. For completeness, we derive here the formula for the CE estimator under label shift. Let $Z_j = |R_j - x_j|^p $ and $m$ denote the number of points in the target distribution.

\begin{align}
    \label{eq:variance}
    \operatorname{Var}\left(\frac{1}{m} \sum_{j=1}^m Z_j \right) &=
    \frac{1}{m^2} \operatorname{Var} \left(\sum_{j=1}^m Z_j \right) \notag \\
    &= \frac{1}{m^2} \left( \sum_{j=1}^m \operatorname{Var} \left( Z_j \right) + \sum_{j \neq i} \operatorname{Cov} \left( Z_j, Z_i \right) \right) \notag \\
    &= \frac{1}{m^2}  \sum_{k=1}^b \left( \sum_{j \in Bin_k} \operatorname{Var} (Z_j)  + \sum_{j \in Bin_k} \sum_{i \in Bin_k, i\neq j} \operatorname{Cov}(Z_j, Z_i) \right) \notag \\
    &= \frac{1}{m^2} \sum_{k=1}^b \left( \frac{m}{b} \operatorname{Var}(Z_j) + \binom{\frac{m}{b}}{2} 2 \operatorname{Cov}(Z_j, Z_i) \right)
\end{align}


\subsection{Empirical evaluation}
In this section, we compare our procedure for obtaining the variance with a Monte Carlo method. We estimate the variance of the CE estimator both on source data, consisting of $n$ samples, and on label-shifted target data containing $m$ samples.
The data is simulated with the following steps:

\begin{enumerate}
    \item We set $p_s(Y=1) = 1/4$ and $p_t(Y=1) = 1/2$, and sample the source and target labels accordingly. 
    \item We sample $p_s(X|Y=1) = p_t(X|Y=1) \sim Beta(alpha=2, beta=1)$
    \item We sample $p_s(X|Y=0) = p_t(X|Y=0) \sim Beta(alpha=2, beta=5)$
\end{enumerate}

\begin{figure}[t]
    \centering
    \subfloat[Source data]{
    \resizebox{0.49\textwidth}{!}{
\begin{tikzpicture}

\definecolor{darkslategray38}{RGB}{38,38,38}
\definecolor{indianred1967981}{RGB}{196,79,81}
\definecolor{lavender234234242}{RGB}{234,234,242}
\definecolor{lightgray204}{RGB}{204,204,204}
\definecolor{steelblue76114176}{RGB}{76,114,176}

\begin{axis}[
axis background/.style={fill=lavender234234242},
axis line style={white},
legend cell align={left},
legend style={
  fill opacity=1,
  draw opacity=1,
  text opacity=1,
  draw=lightgray204,
  fill=lavender234234242
},
scaled x ticks = false,
scaled y ticks = false,
tick align=outside,
tick pos=left,
x grid style={white},
x tick label style={/pgf/number format/fixed},
xlabel=\textcolor{darkslategray38}{Num samples},
xmajorgrids,
xmajorticks=false,
xmajorticks=true,
xmin=-225, xmax=15725,
xtick = {0, 5000, 10000, 15000},
xtick style={color=darkslategray38},
y grid style={white},
ylabel=\textcolor{darkslategray38}{Variance},
ymajorgrids,
ymajorticks=false,
ymajorticks=true,
ymin=-2.56067257082695e-06, ymax=0.000136928456801539,
ytick style={color=darkslategray38}
]
\addplot [line width=1pt, steelblue76114176, mark=*, mark size=3, mark options={solid}]
table {%
500 0.000120044589118948
1000 5.26981477872242e-05
3000 1.89578056458906e-05
5000 1.24317330416781e-05
10000 6.87354523324828e-06
15000 3.77974240064425e-06
};
\addlegendentry{Monte Carlo}
\addplot [line width=1pt, indianred1967981, mark=square*, mark size=3, mark options={solid}]
table {%
500 0.000130588041830068
1000 6.05247839374134e-05
3000 1.92084678419273e-05
5000 1.1775428264348e-05
10000 5.80839972269519e-06
15000 3.78987075796042e-06
};
\addlegendentry{Formula}
\end{axis}

\end{tikzpicture}}
    \label{subfig:variance_synthetic_no_shift}
    }
    \subfloat[Label-shifted target data]{
    \resizebox{0.49\textwidth}{!}{
\begin{tikzpicture}

\definecolor{darkslategray38}{RGB}{38,38,38}
\definecolor{indianred1967981}{RGB}{196,79,81}
\definecolor{lavender234234242}{RGB}{234,234,242}
\definecolor{lightgray204}{RGB}{204,204,204}
\definecolor{steelblue76114176}{RGB}{76,114,176}

\begin{axis}[
axis background/.style={fill=lavender234234242},
axis line style={white},
legend cell align={left},
legend style={
  fill opacity=1,
  draw opacity=1,
  text opacity=1,
  draw=lightgray204,
  fill=lavender234234242
},
scaled x ticks = false,
scaled y ticks = false,
tick align=outside,
tick pos=left,
x grid style={white},
x tick label style={/pgf/number format/fixed},
xlabel=\textcolor{darkslategray38}{Num samples},
xmajorgrids,
xmajorticks=false,
xmajorticks=true,
xmin=-225, xmax=15725,
xtick = {0, 5000, 10000, 15000},
xtick style={color=darkslategray38},
y grid style={white},
ylabel=\textcolor{darkslategray38}{Variance},
ymajorgrids,
ymajorticks=false,
ymajorticks=true,
ymin=-4.09987478348003e-05, ymax=0.00114865322884928,
ytick style={color=darkslategray38}
]
\addplot [line width=1pt, steelblue76114176, mark=*, mark size=3, mark options={solid}]
table {%
500 0.001094578139
1000 0.000303520004
3000 8.2442256e-05
5000 6.8844156e-05
10000 2.76416e-05
15000 2.3105011e-05
};
\addlegendentry{Monte Carlo}
\addplot [line width=1pt, indianred1967981, mark=square*, mark size=3, mark options={solid}]
table {%
500 0.000756715215818856
1000 0.000310289597912418
3000 7.56010142650384e-05
5000 4.62137188413483e-05
10000 2.67989277875617e-05
15000 1.3076342014476e-05
};
\addlegendentry{Formula}
\end{axis}

\end{tikzpicture}}
    \label{subfig:variance_synthetic_label_shift}
    }
    \caption{Comparison of our variance procedure with a Monte Carlo method for various sample sizes. }
    \label{fig:variance_cmp_monte_carlo}
\end{figure}
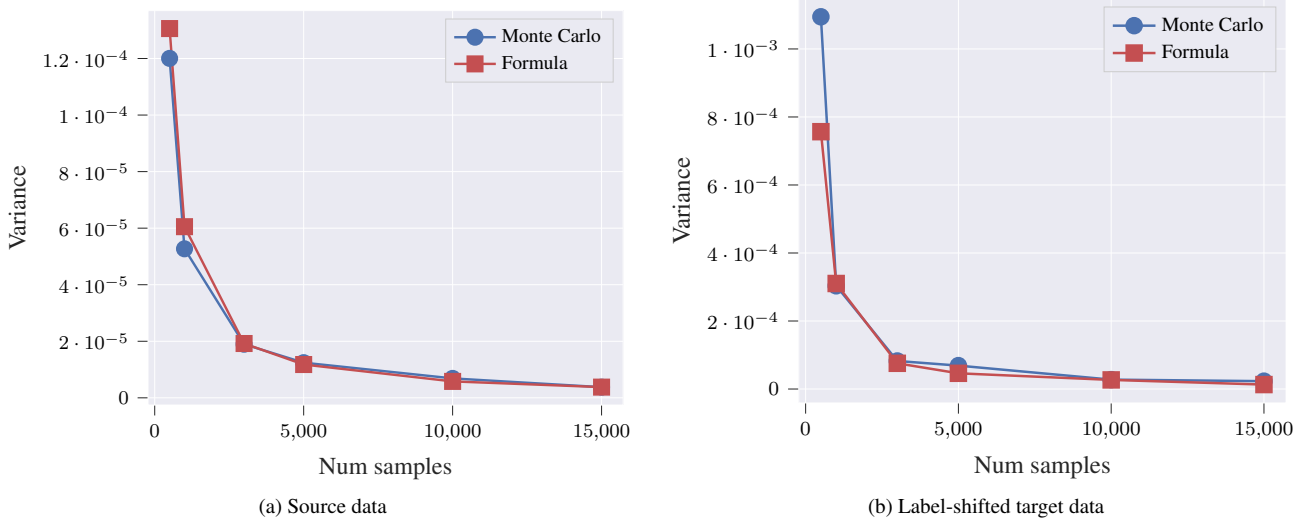

In Figure~\ref{fig:variance_cmp_monte_carlo} we show a comparison of our variance procedure, given by Equation~\ref{eq:variance} and Algorithm~\ref{alg:calculate_variance}, and a Monte Carlo method. Within each Monte Carlo simulation, we sample new points according to the described procedure and calculate the calibration error using our CE estimator. Subsequently, we determine the sample variance across 100 simulations. The results demonstrate a notable similarity in the variances obtained through both methods. This shows that our proposed method effectively approximates the variance, aligning closely with the values obtained with the widely-used Monte Carlo simulation. Importantly, our method does not suffer from the computational costs associated with a large number of simulations required in the Monte Carlo method.

\section{Experiments}\label{app:experiments}
In this section, we include more details about the used datasets and training procedures. We also report additional experiments to evaluate the performance of our proposed method using different importance weight estimators.

\subsection{Details about the datasets}
We report statistics for all datasets in Table~\ref{table:datasets-info}.

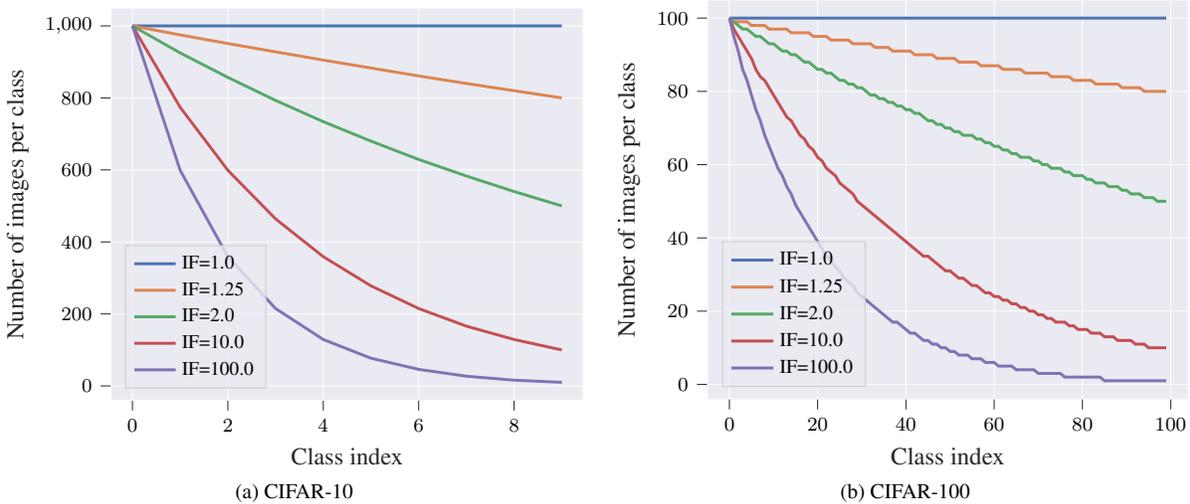
\begin{figure}[t]
    \centering
    \subfloat[CIFAR-10]{
    \resizebox{0.45\textwidth}{!}{
\begin{tikzpicture}

\definecolor{darkslategray38}{RGB}{38,38,38}
\definecolor{indianred1967882}{RGB}{196,78,82}
\definecolor{lavender234234242}{RGB}{234,234,242}
\definecolor{lightgray204}{RGB}{204,204,204}
\definecolor{lightslategray129114179}{RGB}{129,114,179}
\definecolor{mediumseagreen85168104}{RGB}{85,168,104}
\definecolor{peru22113282}{RGB}{221,132,82}
\definecolor{steelblue76114176}{RGB}{76,114,176}

\begin{axis}[
axis background/.style={fill=lavender234234242},
axis line style={white},
legend cell align={left},
legend style={
  fill opacity=0.8,
  draw opacity=1,
  text opacity=1,
  at={(0.03,0.03)},
  anchor=south west,
  draw=lightgray204,
  fill=lavender234234242
},
scaled x ticks = false,
scaled y ticks = false,
tick align=outside,
tick pos=left,
x grid style={white},
x tick label style={/pgf/number format/fixed},
xlabel=\textcolor{darkslategray38}{Class index},
xmajorgrids,
xmajorticks=false,
xmajorticks=true,
xmin=-0.45, xmax=9.45,
xtick style={color=darkslategray38},
y grid style={white},
y tick label style={/pgf/number format/.cd, fixed},
ylabel=\textcolor{darkslategray38}{Number of images per class},
ymajorgrids,
ymajorticks=false,
ymajorticks=true,
ymin=-39.5, ymax=1049.5,
ytick style={color=darkslategray38}
]
\addplot [very thick, steelblue76114176]
table {%
0 1000
1 1000
2 1000
3 1000
4 1000
5 1000
6 1000
7 1000
8 1000
9 1000
};
\addlegendentry{IF=1.0}
\addplot [very thick, peru22113282]
table {%
0 1000
1 975
2 951
3 928
4 905
5 883
6 861
7 840
8 820
9 800
};
\addlegendentry{IF=1.25}
\addplot [very thick, mediumseagreen85168104]
table {%
0 1000
1 925
2 857
3 793
4 734
5 680
6 629
7 583
8 540
9 500
};
\addlegendentry{IF=2.0}
\addplot [very thick, indianred1967882]
table {%
0 1000
1 774
2 599
3 464
4 359
5 278
6 215
7 166
8 129
9 100
};
\addlegendentry{IF=10.0}
\addplot [very thick, lightslategray129114179]
table {%
0 1000
1 599
2 359
3 215
4 129
5 77
6 46
7 27
8 16
9 10
};
\addlegendentry{IF=100.0}
\end{axis}

\end{tikzpicture}}
    \label{subfig:label_distributions_cifar10}
    }
    \subfloat[CIFAR-100]{
    \resizebox{0.45\textwidth}{!}{
\begin{tikzpicture}

\definecolor{darkslategray38}{RGB}{38,38,38}
\definecolor{indianred1967882}{RGB}{196,78,82}
\definecolor{lavender234234242}{RGB}{234,234,242}
\definecolor{lightgray204}{RGB}{204,204,204}
\definecolor{lightslategray129114179}{RGB}{129,114,179}
\definecolor{mediumseagreen85168104}{RGB}{85,168,104}
\definecolor{peru22113282}{RGB}{221,132,82}
\definecolor{steelblue76114176}{RGB}{76,114,176}

\begin{axis}[
axis background/.style={fill=lavender234234242},
axis line style={white},
legend cell align={left},
legend style={
  fill opacity=0.8,
  draw opacity=1,
  text opacity=1,
  at={(0.03,0.03)},
  anchor=south west,
  draw=lightgray204,
  fill=lavender234234242
},
scaled x ticks = false,
scaled y ticks = false,
tick align=outside,
tick pos=left,
x grid style={white},
x tick label style={/pgf/number format/fixed},
xlabel=\textcolor{darkslategray38}{Class index},
xmajorgrids,
xmajorticks=false,
xmajorticks=true,
xmin=-4.95, xmax=103.95,
xtick style={color=darkslategray38},
y grid style={white},
y tick label style={/pgf/number format/.cd, fixed},
ylabel=\textcolor{darkslategray38}{Number of images per class},
ymajorgrids,
ymajorticks=false,
ymajorticks=true,
ymin=-3.95, ymax=104.95,
ytick style={color=darkslategray38}
]
\addplot [very thick, steelblue76114176]
table {%
0 100
1 100
2 100
3 100
4 100
5 100
6 100
7 100
8 100
9 100
10 100
11 100
12 100
13 100
14 100
15 100
16 100
17 100
18 100
19 100
20 100
21 100
22 100
23 100
24 100
25 100
26 100
27 100
28 100
29 100
30 100
31 100
32 100
33 100
34 100
35 100
36 100
37 100
38 100
39 100
40 100
41 100
42 100
43 100
44 100
45 100
46 100
47 100
48 100
49 100
50 100
51 100
52 100
53 100
54 100
55 100
56 100
57 100
58 100
59 100
60 100
61 100
62 100
63 100
64 100
65 100
66 100
67 100
68 100
69 100
70 100
71 100
72 100
73 100
74 100
75 100
76 100
77 100
78 100
79 100
80 100
81 100
82 100
83 100
84 100
85 100
86 100
87 100
88 100
89 100
90 100
91 100
92 100
93 100
94 100
95 100
96 100
97 100
98 100
99 100
};
\addlegendentry{IF=1.0}
\addplot [very thick, peru22113282]
table {%
0 100
1 99
2 99
3 99
4 99
5 98
6 98
7 98
8 98
9 97
10 97
11 97
12 97
13 97
14 96
15 96
16 96
17 96
18 96
19 95
20 95
21 95
22 95
23 94
24 94
25 94
26 94
27 94
28 93
29 93
30 93
31 93
32 93
33 92
34 92
35 92
36 92
37 91
38 91
39 91
40 91
41 91
42 90
43 90
44 90
45 90
46 90
47 89
48 89
49 89
50 89
51 89
52 88
53 88
54 88
55 88
56 88
57 87
58 87
59 87
60 87
61 87
62 86
63 86
64 86
65 86
66 86
67 85
68 85
69 85
70 85
71 85
72 85
73 84
74 84
75 84
76 84
77 84
78 83
79 83
80 83
81 83
82 83
83 82
84 82
85 82
86 82
87 82
88 82
89 81
90 81
91 81
92 81
93 81
94 80
95 80
96 80
97 80
98 80
99 80
};
\addlegendentry{IF=1.25}
\addplot [very thick, mediumseagreen85168104]
table {%
0 100
1 99
2 98
3 97
4 97
5 96
6 95
7 95
8 94
9 93
10 93
11 92
12 91
13 91
14 90
15 90
16 89
17 88
18 88
19 87
20 86
21 86
22 85
23 85
24 84
25 83
26 83
27 82
28 82
29 81
30 81
31 80
32 79
33 79
34 78
35 78
36 77
37 77
38 76
39 76
40 75
41 75
42 74
43 74
44 73
45 72
46 72
47 71
48 71
49 70
50 70
51 69
52 69
53 68
54 68
55 68
56 67
57 67
58 66
59 66
60 65
61 65
62 64
63 64
64 63
65 63
66 62
67 62
68 62
69 61
70 61
71 60
72 60
73 59
74 59
75 59
76 58
77 58
78 57
79 57
80 57
81 56
82 56
83 55
84 55
85 55
86 54
87 54
88 54
89 53
90 53
91 52
92 52
93 52
94 51
95 51
96 51
97 50
98 50
99 50
};
\addlegendentry{IF=2.0}
\addplot [very thick, indianred1967882]
table {%
0 100
1 97
2 95
3 93
4 91
5 89
6 86
7 84
8 83
9 81
10 79
11 77
12 75
13 73
14 72
15 70
16 68
17 67
18 65
19 64
20 62
21 61
22 59
23 58
24 57
25 55
26 54
27 53
28 52
29 50
30 49
31 48
32 47
33 46
34 45
35 44
36 43
37 42
38 41
39 40
40 39
41 38
42 37
43 36
44 35
45 35
46 34
47 33
48 32
49 31
50 31
51 30
52 29
53 29
54 28
55 27
56 27
57 26
58 25
59 25
60 24
61 24
62 23
63 23
64 22
65 22
66 21
67 21
68 20
69 20
70 19
71 19
72 18
73 18
74 17
75 17
76 17
77 16
78 16
79 15
80 15
81 15
82 14
83 14
84 14
85 13
86 13
87 13
88 12
89 12
90 12
91 12
92 11
93 11
94 11
95 10
96 10
97 10
98 10
99 10
};
\addlegendentry{IF=10.0}
\addplot [very thick, lightslategray129114179]
table {%
0 100
1 95
2 91
3 86
4 83
5 79
6 75
7 72
8 68
9 65
10 62
11 59
12 57
13 54
14 52
15 49
16 47
17 45
18 43
19 41
20 39
21 37
22 35
23 34
24 32
25 31
26 29
27 28
28 27
29 25
30 24
31 23
32 22
33 21
34 20
35 19
36 18
37 17
38 17
39 16
40 15
41 14
42 14
43 13
44 12
45 12
46 11
47 11
48 10
49 10
50 9
51 9
52 8
53 8
54 8
55 7
56 7
57 7
58 6
59 6
60 6
61 5
62 5
63 5
64 5
65 4
66 4
67 4
68 4
69 4
70 3
71 3
72 3
73 3
74 3
75 3
76 2
77 2
78 2
79 2
80 2
81 2
82 2
83 2
84 2
85 1
86 1
87 1
88 1
89 1
90 1
91 1
92 1
93 1
94 1
95 1
96 1
97 1
98 1
99 1
};
\addlegendentry{IF=100.0}
\end{axis}

\end{tikzpicture}}
    \label{subfig:label_distributions_cifar100}
    }
    \caption{Number of target samples per class in simulated long-tail CIFAR-10/100 datasets with different imbalance factors (IF).}
    \label{fig:label_distribution_cifars}
\end{figure}

\textbf{CIFAR-10/100 \cite{krizhevsky2009learning, cao2019learning}.}
Using the CIFAR datasets we examine two different types of label shift. In \textbf{Task A} the source distribution is long-tailed, whereas the target is uniform. In \textbf{Task B} the source distribution is uniform, and the target is long-tailed. The CIFAR10/100 Long-Tail datasets are simulated from CIFAR10/100, respectively, with different imbalance factors (IF). The IF controls the ratio between the number of samples in the most frequent and the least frequent class. For example, an imbalance factor of 10 indicates that the least frequent class appears 10 times less than the most frequent one. 
In \textbf{Task A} we keep the target distribution unchanged (i.e., balanced across classes), and we resample the source distribution with various IF. In the main paper, we presented a setting with source $\operatorname{IF}=10$ in Table~\ref{table:cifarlt-imagenetlt}. Additional results using different imbalance factors induced on the source distribution are given in Tables~\ref{table:cifarlt-imagenetlt-all-estimators-0.1} -- \ref{table:cifarlt-imagenetlt-all-estimators-0.5-p=1}.
In \textbf{Task B} the models are trained on the original (balanced) CIFAR datasets, and in Table~\ref{table:balanced-cifar} we report the performance of our CE estimator on label-shifted target distribution with various imbalance factors. In Figure~\ref{fig:label_distribution_cifars} we show the number of target images per class on the long-tailed CIFAR-10/100 with imbalance factors ranging from 1.25 to 100. 

\begin{table*}[t]
\centering
\resizebox{1.0\textwidth}{!}{
\begin{tabular}{cccccc} \toprule
   {Dataset} & {Modality} & {Num. classes} & {Training num. samples} & {Validation num. samples} & {Testing num. samples} \\ \midrule
   CIFAR10 & Images & 10 & 40,000 & 10,000 & 10,000 \\
   CIFAR100 & Images & 100 & 40,000 & 10,000 & 10,000 \\
   ImageNet Long-Tail & Images & 1000 & 108,676 & 27,170 & 50,000 \\
   Camelyon17 & Images & 2 & 302436 & 33560 & -- \\
   iWildCam & Images & 182 & 129809 & 7314 & 8154 \\
   Amazon & Text & 5 & 245502 & 46950 & 46950 \\ \bottomrule
\end{tabular}
}
\caption{Statistics for all datasets used in the paper. Note that we report the original number of classes and samples of the datasets we use.}
\label{table:datasets-info}
\end{table*}

\textbf{ImageNet Long-Tail \cite{liu2019largescale}.} We use this dataset to verify that our findings also hold on large-scale datasets. The dataset is obtained from ImageNet, such that the training and the validation set (source distribution) exhibit a long-tail distribution, while the test set (target distribution) is uniform across classes. To ensure that the training and validation set follow the same distribution, we merge the original training and validation set, and then divide them into new train and validation dataset using stratified sampling as per Chen \etal \cite{chen2023transfer} -- we keep 20\% of all samples as a holdout validation set. Finally, we evaluate on all classes with a frequency between 20 and 50, such that we exclude both infrequent classes (for which we cannot obtain reliable importance weights), and head classes (which exhibit a high frequency and are therefore easier to learn). The test dataset (target distribution) features 619 classes and 30950 samples in total, while the validation set (source distribution) contains 14442 samples.

\textbf{Camelyon17 \cite{bandi2018detection}} consists of $96 \times 96$ whole-side images (WSI) of breast-cancer metastases in lymph node sections collected from hospitals in the Netherlands. In each WSI, the tumor regions are annotated manually by pathologists. The labels indicate whether the central $32 \times 32$ region contains a tumor. As Camelyon17 contains only a training and validation set -- drawn from the same (source) distribution -- both of which are balanced across the positive and negative class, we perform the following: (i) We use the validation set as testing dataset, which we convert to our desired target distribution by resampling the positive class; (ii) From the training dataset, we allocate a validation dataset with the same size as the testing dataset. Then, we subsample the validation dataset the same way as we subsample the training dataset, so that both are effectively drawn from the same (source) distribution (e.g. used in the ablation studies in Section~4.3).

\textbf{iWildCam \cite{beery2021iwildcam}} consists of images obtained from animal camera traps -- heat or motion-activated static cameras placed in the wild -- which are set in countries in different parts of the world. The label of each image is one of the 182 animal species.
The training and validation set feature the same long-tail distribution among classes. As testing data, we use the original test dataset and we perform the following: (i) we keep only the 20 most frequent classes; (ii) we resample the dataset such that classes follow a uniform distribution (thus representing the target label-shifted distribution) -- we impose a minimal frequency of 84 samples per testing class. Therefore, the number of samples in the test dataset (target distribution) is 1680, while the validation set (source distribution) contains 6003 samples.

\textbf{Amazon \cite{ni2019justifying}} consists of texts which represent user reviews, while the label is 1-out-of-5 score of the review. The training and validation set -- source distribution -- follow the same long-tail distribution among classes. As the testing dataset is also long-tail (as per the training dataset), we resample each class based on the frequency of the least frequent class, yielding a test set following a uniform distribution of classes, representing the target, where each class appears 527 times. Therefore, the number of samples in the test dataset (target distribution) is 2860.

\subsection{Implementation details}
We conduct all experiments on consumer-grade GPUs, that is, all experiments can be conducted on a single Nvidia 3090. We use PyTorch \cite{paszke2019pytorch} for all deep-learning-based implementations. Below we provide further information about the training procedure for each of the datasets, along with implementation details of the weight estimators we use.

\textbf{CIFAR10/100 (Long-Tail).}  We keep the same training procedure for both CIFAR10/100 and their long-tail variants. Namely, we train all models with stochastic gradient descent (SGD) for 200 epochs, with a peak learning rate of 0.1, linearly warmed up for the first 10\% of the training, and then decreased to 0.0 until the end. We apply weight decay of 0.0005, and clip the gradients when their norm exceeds 5.0. During training, we augment the images by applying random horizontal flips.

\textbf{ImageNet Long-Tail.} We train all models with AdamW \cite{loshchilov2017decoupled} for 100 epochs, with a peak learning rate of 0.0005, linearly warmed up for the first 10\% of the training, and then decreased to 0.0 until the end. We apply weight decay of 0.001, and clip the gradients when their norm exceeds 5.0. During training, we obtain a random crop of the image with size $224 \times 224$, perform horizontal flipping, and apply color jitter with parameters: brightness=(0.6, 1.4), contrast=(0.6, 1.4), saturation=(0.6, 1.4). During testing, we obtain a single crop with size $224 \times 224$ from the center of the image.

\textbf{WILDS datasets (Camelyon17, iWildCam and Amazon)} We keep the same training procedure across all WILDS datasets, with the only difference across datasets being the data augmentation and the used models. Namely, we train all models with AdamW for 10 epochs, with a peak learning rate of 0.0005, linearly warmed up for the first 10\% of the training and then decreased to 0.0 until the end. We apply weight decay of 0.001, and clip the gradients when their norm exceeds 5.0. During training, for models trained on Amazon we do not apply any data augmentation on the input text, while for models trained on Camelyon17 and iWildCam we apply the same data augmentation as with the ImageNet Long-Tail based models. During testing, we obtain a single crop with size $224 \times 224$ from the center of the image. For all ImageNet pre-trained models we use Timm \cite{wightman2019pytorch} (Camelyon17 and iWildCam), while for all pre-trained language models we use HuggingFace transformers \cite{wolf2019huggingface}.
On Camelyon17 and iWildCam, we train diverse set of transformer based-models which are pre-trained on ImageNet: ResNet50, ViT-Large, ViT-Large (with input resultion of 384), and Swin-Large. On Amazon, we train different transformer-based language models: BERT (bert-base-uncased), D-BERT (distilbert-base-uncased), RoBERTa (roberta-base) and D-RoBERTa (distilroberta-base).

\textbf{Weight estimators.} Our proposed method relies on estimating importance weights using techniques from the unsupervised domain adaptation literature. Most of the weight estimators (RLLS \cite{azizzadenesheli2019regularized}, BBSL \cite{lipton2018detecting}, EM-BCTS \cite{saerens2002adjusting}) that we use are implemented in \url{https://github.com/kundajelab/abstention}. For the ELSA \cite{tian2023elsa} method, we used the original implementation provided by the authors. In some of our settings, we detected issues with the EM-BCTS and ELSA methods, prompting us to set a minimal value of the confidence scores to $1 \times 10^{-15}$ for EM-BCTS and $1 \times 10^{-3}$ for ELSA, in order to get a reasonable estimate of the weights. We also encountered issues with the BBSL method on the iWildCam dataset, due to the source distribution containing 0-frequency classes. RLLS consistently delivered the most accurate and stable weight estimations, thus, we report our main results using this method.
Note that in certain experiments, some of the importance weight estimation methods (e.g. BBSL) yield poor estimates, resulting in abnormal values for the calibration error. However, addressing these issues is beyond the scope of this paper, as they are specific to the weight estimation methods, and not with our CE estimator.

\subsection{Effect of different importance weight estimators}
In this section, we report additional experiments to assess the effectiveness of our proposed approach using various importance weight estimation methods: RLLS, ELSA, EM-BCTS, and BBSL. 


\subsubsection{Experiments on natural image datasets}

In Tables~\ref{table:cifarlt-imagenetlt-all-estimators-0.1} -- \ref{table:cifarlt-imagenetlt-all-estimators-0.5-p=1} we report accuracy, ground truth CE (using labels) and estimated CE using different importance weight estimators. The models are trained on CIFAR-10/100-LT. Each table corresponds to different IF imposed on the source distribution, and we report CE with different $L_p$ norms -- $L_1$ or $L_2$. For all experiments, the target distribution is uniform. The subscripts $s$ and $t$ denote the source and target distributions, respectively.

When encountering a less severe label shift (Table~\ref{table:cifarlt-imagenetlt-all-estimators-0.2} ($\operatorname{IF}=5$) and Table~\ref{table:cifarlt-imagenetlt-all-estimators-0.5} ($\operatorname{IF}=2$)), we observe a comparable performance across all weight estimators. However, under more pronounced label shift (Table \ref{table:cifarlt-imagenetlt-all-estimators-0.1} ($\operatorname{IF}=10$)), in several settings we encounter issues with ELSA, EM-BCTS and BBSL methods, resulting in abnormal CE values.
In contrast, the RLLS method yields stable and reliable values across all settings. The CE estimates obtained using RLLS often closely align with those of the CE estimator that utilizes ground truth weights, denoted as $\hat{\bo}^*$. 


\begin{table*}[t]
\centering
\resizebox{1.0\textwidth}{!}{
\begin{tabular}{ccccccc>{\columncolor{method}}c>{\columncolor{method}}c>{\columncolor{method}}c>{\columncolor{method}}c} \toprule

   {Dataset} & {Model} & {Accuracy$_s$} & {Accuracy$_t$} & {$\operatorname{CE}_s$} & {$\operatorname{CE}_t$} & {$\widehat{\operatorname{CE}}_t(\hat{\bo}^*)$ } & \makecell{\textbf{$\widehat{\operatorname{CE}}_t (\hat{\bo})$} \\ $\text{RLLS}$} & 
   \makecell{\textbf{$\widehat{\operatorname{CE}}_t (\hat{\bo})$} \\ $\text{ELSA}$} & 
   \makecell{\textbf{$\widehat{\operatorname{CE}}_t (\hat{\bo})$} \\ $\text{EM-BCTS}$} & 
   \makecell{\textbf{$\widehat{\operatorname{CE}}_t (\hat{\bo})$} \\ $\text{BBSL}$} \\ 
   
   \midrule
    \multirow{4}{*}{CIFAR-10-LT}
 & ResNet-20  & 83.10 & 78.24 & 0.81 & 0.88  & 0.91  & 0.89  & 0.89  & 0.84  & 0.89  \\
 & ResNet-32  & 85.48 & 80.74 & 0.90 & 1.04  & 1.16  & 1.21  & 1.20  & 1.13  & 1.21  \\
 & ResNet-56  & 85.38 & 81.71 & 0.98 & 1.12  & 1.16  & 1.16  & 1.15  & 1.10  & 1.16  \\
 & ResNet-110  & 84.94 & 81.29 & 1.00 & 1.19  & 1.22  & 1.22  & 1.22  & 1.16  & 1.22  \\
    \midrule
    \multirow{4}{*}{CIFAR-100-LT} 
 & ResNet-20  & 52.24 & 44.81 & 0.61 & 0.66  & 0.66  & 0.64  & 0.96  & 0.64  & 1344.57  \\
 & ResNet-32  & 53.48 & 47.73 & 0.66 & 0.71  & 0.71  & 0.70  & 0.83  & 0.70  & 4.69  \\
 & ResNet-56  & 54.21 & 47.18 & 0.66 & 0.72  & 0.72  & 0.71  & 10.44  & 0.70  & 118.43  \\
 & ResNet-110  & 56.58 & 49.78 & 0.69 & 0.73  & 0.73  & 0.72  & 0.77  & 0.71  & 0.83  \\ \bottomrule
\end{tabular}
}
\caption{Comparison of different importance weight estimators. The source is obtained with an \textbf{$\operatorname{IF}=10$}. We measure $L_2$ CE. The abnormal values on CIFAR-100-LT with BBSL and ELSA are due to issues of the weight estimators in this setting. }
\label{table:cifarlt-imagenetlt-all-estimators-0.1}
\end{table*}

\begin{table*}[t]
\centering
\resizebox{1.0\textwidth}{!}{
\begin{tabular}{ccccccc>{\columncolor{method}}c>{\columncolor{method}}c>{\columncolor{method}}c>{\columncolor{method}}c} \toprule
   {Dataset} & {Model} & {Accuracy$_s$} &  {Accuracy$_t$} & {$\operatorname{CE}_s$} & {$\operatorname{CE}_t$} & {$\widehat{\operatorname{CE}}_t(\hat{\bo}^*)$ } & \makecell{\textbf{$\widehat{\operatorname{CE}}_t (\hat{\bo})$} \\ $\text{RLLS}$} & 
   \makecell{\textbf{$\widehat{\operatorname{CE}}_t (\hat{\bo})$} \\ $\text{ELSA}$} & 
   \makecell{\textbf{$\widehat{\operatorname{CE}}_t (\hat{\bo})$} \\ $\text{EM-BCTS}$} & 
   \makecell{\textbf{$\widehat{\operatorname{CE}}_t (\hat{\bo})$} \\ $\text{BBSL}$} \\ 
   \midrule
    \multirow{4}{*}{CIFAR-10-LT}
 & ResNet-20  & 84.77 & 82.77 & 0.71 & 0.84  & 0.90  & 0.90  & 0.91  & 0.88  & 0.90  \\
 & ResNet-32  & 86.68 & 83.47 & 0.79 & 1.03  & 1.02  & 1.03  & 1.03  & 1.00  & 1.03  \\
 & ResNet-56  & 86.03 & 84.17 & 0.90 & 1.19  & 1.18  & 1.17  & 1.17  & 1.15  & 1.17  \\
 & ResNet-110  & 86.44 & 85.04 & 0.96 & 1.32  & 1.35  & 1.35  & 1.35  & 1.33  & 1.35  \\
    \midrule
    \multirow{4}{*}{CIFAR-100-LT} 
 & ResNet-20  & 52.92 & 50.39 & 0.62 & 0.65  & 0.65  & 0.66  & 0.67  & 0.64  & 0.69  \\
 & ResNet-32  & 55.52 & 50.64 & 0.69 & 0.71  & 0.71  & 0.72  & 0.73  & 0.70  & 0.74  \\
 & ResNet-56  & 56.59 & 53.33 & 0.70 & 0.73  & 0.73  & 0.74  & 0.76  & 0.72  & 0.76  \\
 & ResNet-110  & 57.26 & 54.16 & 0.71 & 0.74  & 0.74  & 0.74  & 0.75  & 0.73  & 0.76  \\ \bottomrule
\end{tabular}
}
\caption{Comparison of different importance weight estimators. The source is obtained with an \textbf{$\operatorname{IF}=5$}. We measure $L_2$ CE.}
\label{table:cifarlt-imagenetlt-all-estimators-0.2}
\end{table*}

\begin{table*}[t]
\centering
\resizebox{1.0\textwidth}{!}{
\begin{tabular}{ccccccc>{\columncolor{method}}c>{\columncolor{method}}c>{\columncolor{method}}c>{\columncolor{method}}c} \toprule
   {Dataset} & {Model} & {Accuracy$_s$} &  {Accuracy$_t$} & {$\operatorname{CE}_s$} & {$\operatorname{CE}_t$} & {$\widehat{\operatorname{CE}}_t(\hat{\bo}^*)$ } & 
   \makecell{\textbf{$\widehat{\operatorname{CE}}_t (\hat{\bo})$} \\ $\text{RLLS}$} & 
   \makecell{\textbf{$\widehat{\operatorname{CE}}_t (\hat{\bo})$} \\ $\text{ELSA}$} & 
   \makecell{\textbf{$\widehat{\operatorname{CE}}_t (\hat{\bo})$} \\ $\text{EM-BCTS}$} & 
   \makecell{\textbf{$\widehat{\operatorname{CE}}_t (\hat{\bo})$} \\ $\text{BBSL}$} \\ 
   \midrule
    \multirow{4}{*}{CIFAR-10-LT}
 & ResNet-20  & 86.74 & 85.78 & 0.51 & 0.91  & 0.95  & 0.95  & 0.96  & 0.95  & 0.95  \\
 & ResNet-32  & 86.94 & 86.82 & 0.55 & 1.11  & 1.11  & 1.11  & 1.10  & 1.11  & 1.11  \\
 & ResNet-56  & 88.41 & 87.93 & 0.78 & 1.38  & 1.37  & 1.38  & 1.38  & 1.37  & 1.38  \\
 & ResNet-110  & 88.33 & 87.51 & 0.76 & 1.33  & 1.34  & 1.37  & 1.36  & 1.33  & 1.37  \\
    \midrule
    \multirow{4}{*}{CIFAR-100-LT} 
 & ResNet-20  & 56.81 & 56.71 & 0.60 & 0.61  & 0.61  & 0.62  & 0.62  & 0.61  & 1.23  \\
 & ResNet-32  & 58.43 & 58.59 & 0.70 & 0.71  & 0.71  & 0.71  & 0.71  & 0.70  & 0.72  \\
 & ResNet-56  & 60.42 & 59.96 & 0.74 & 0.74  & 0.74  & 0.75  & 0.75  & 0.74  & 0.75  \\
 & ResNet-110  & 62.88 & 61.99 & 0.75 & 0.76  & 0.75  & 0.76  & 0.76  & 0.75  & 0.76  \\ \bottomrule
\end{tabular}
}
\caption{Comparison of different importance weight estimators. The source is obtained with an \textbf{$\operatorname{IF}=2$}. We measure $L_2$ CE.}
\label{table:cifarlt-imagenetlt-all-estimators-0.5}
\end{table*}

\begin{table*}[t]
\centering
\resizebox{1.0\textwidth}{!}{
\begin{tabular}{ccccccc>{\columncolor{method}}c>{\columncolor{method}}c>{\columncolor{method}}c>{\columncolor{method}}c} \toprule
   {Dataset} & {Model} & {Accuracy$_s$} &  {Accuracy$_t$} & {$\operatorname{CE}_s$} & {$\operatorname{CE}_t$} & {$\widehat{\operatorname{CE}}_t(\hat{\bo}^*)$ } & 
   \makecell{\textbf{$\widehat{\operatorname{CE}}_t (\hat{\bo})$} \\ $\text{RLLS}$} & 
   \makecell{\textbf{$\widehat{\operatorname{CE}}_t (\hat{\bo})$} \\ $\text{ELSA}$} & 
   \makecell{\textbf{$\widehat{\operatorname{CE}}_t (\hat{\bo})$} \\ $\text{EM-BCTS}$} & 
   \makecell{\textbf{$\widehat{\operatorname{CE}}_t (\hat{\bo})$} \\ $\text{BBSL}$} \\ 
   \midrule
    \multirow{4}{*}{CIFAR-10-LT}
 & ResNet-20  & 83.10 & 78.24 & 3.16 & 3.58  & 3.50  & 3.46  & 3.44  & 3.54  & 3.46  \\
 & ResNet-32  & 85.48 & 80.74 & 3.26 & 3.76  & 3.86  & 3.91  & 3.88  & 4.04  & 3.91  \\
 & ResNet-56  & 85.38 & 81.71 & 3.39 & 3.80  & 3.88  & 3.84  & 3.81  & 3.88  & 3.84  \\
 & ResNet-110  & 84.94 & 81.29 & 3.43 & 3.87  & 3.94  & 3.95  & 3.92  & 4.03  & 3.95  \\
    \midrule
    \multirow{4}{*}{CIFAR-100-LT} 
 & ResNet-20  & 52.24 & 44.81 & 1.46 & 1.57  & 1.57  & 1.47  & 1.99  & 1.53  & 50.91  \\
 & ResNet-32  & 53.48 & 47.73 & 1.52 & 1.62  & 1.63  & 1.54  & 1.86  & 1.59  & 4.00  \\
 & ResNet-56  & 54.21 & 47.18 & 1.51 & 1.64  & 1.64  & 1.56  & 5.10  & 1.60  & 15.05  \\
 & ResNet-110  & 56.58 & 49.78 & 1.54 & 1.64  & 1.64  & 1.57  & 1.69  & 1.61  & 1.75  \\ \bottomrule
\end{tabular}
}
\caption{Comparison of different importance weight estimators. The source is obtained with an \textbf{$\operatorname{IF}=10$}. We measure $L_1$ CE.}
\label{table:cifarlt-imagenetlt-all-estimators-0.1-p=1}
\end{table*}

\begin{table*}[t]
\centering
\resizebox{1.0\textwidth}{!}{
\begin{tabular}{ccccccc>{\columncolor{method}}c>{\columncolor{method}}c>{\columncolor{method}}c>{\columncolor{method}}c} \toprule
   {Dataset} & {Model} & {Accuracy$_s$} &  {Accuracy$_t$} & {$\operatorname{CE}_s$} & {$\operatorname{CE}_t$} & {$\widehat{\operatorname{CE}}_t(\hat{\bo}^*)$ } & 
   \makecell{\textbf{$\widehat{\operatorname{CE}}_t (\hat{\bo})$} \\ $\text{RLLS}$} & 
   \makecell{\textbf{$\widehat{\operatorname{CE}}_t (\hat{\bo})$} \\ $\text{ELSA}$} & 
   \makecell{\textbf{$\widehat{\operatorname{CE}}_t (\hat{\bo})$} \\ $\text{EM-BCTS}$} & 
   \makecell{\textbf{$\widehat{\operatorname{CE}}_t (\hat{\bo})$} \\ $\text{BBSL}$} \\ 
   \midrule
    \multirow{4}{*}{CIFAR-10-LT}
 & ResNet-20  & 84.77 & 82.77 & 2.85 & 2.99  & 3.11  & 3.17  & 3.20  & 3.23  & 3.17  \\
 & ResNet-32  & 86.68 & 83.47 & 2.87 & 3.32  & 3.17  & 3.22  & 3.21  & 3.29  & 3.22  \\
 & ResNet-56  & 86.03 & 84.17 & 3.15 & 3.68  & 3.64  & 3.55  & 3.56  & 3.63  & 3.55  \\
 & ResNet-110  & 86.44 & 85.04 & 3.22 & 3.81  & 3.86  & 3.85  & 3.86  & 3.87  & 3.85  \\
    \midrule
    \multirow{4}{*}{CIFAR-100-LT} 
 & ResNet-20  & 52.92 & 50.39 & 1.49 & 1.55  & 1.55  & 1.54  & 1.58  & 1.53  & 1.62  \\
 & ResNet-32  & 55.52 & 50.64 & 1.56 & 1.62  & 1.62  & 1.60  & 1.64  & 1.60  & 1.64  \\
 & ResNet-56  & 56.59 & 53.33 & 1.58 & 1.63  & 1.64  & 1.62  & 1.67  & 1.61  & 1.66  \\
 & ResNet-110  & 57.26 & 54.16 & 1.58 & 1.64  & 1.64  & 1.62  & 1.64  & 1.61  & 1.65  \\ \bottomrule
\end{tabular}
}
\caption{Comparison of different importance weight estimators. The source is obtained with an \textbf{$\operatorname{IF}=5$}. We measure $L_1$ CE.}
\label{table:cifarlt-imagenetlt-all-estimators-0.2-p=1}
\end{table*}

\begin{table*}[t]
\centering
\resizebox{1.0\textwidth}{!}{
\begin{tabular}{ccccccc>{\columncolor{method}}c>{\columncolor{method}}c>{\columncolor{method}}c>{\columncolor{method}}c} \toprule
   {Dataset} & {Model} & {Accuracy$_s$} &  {Accuracy$_t$} & {$\operatorname{CE}_s$} & {$\operatorname{CE}_t$} & {$\widehat{\operatorname{CE}}_t(\hat{\bo}^*)$ } & 
   \makecell{\textbf{$\widehat{\operatorname{CE}}_t (\hat{\bo})$} \\ $\text{RLLS}$} & 
   \makecell{\textbf{$\widehat{\operatorname{CE}}_t (\hat{\bo})$} \\ $\text{ELSA}$} & 
   \makecell{\textbf{$\widehat{\operatorname{CE}}_t (\hat{\bo})$} \\ $\text{EM-BCTS}$} & 
   \makecell{\textbf{$\widehat{\operatorname{CE}}_t (\hat{\bo})$} \\ $\text{BBSL}$} \\ 
   \midrule
    \multirow{4}{*}{CIFAR-10-LT}
 & ResNet-20  & 86.74 & 85.78 & 2.23 & 2.89  & 2.96  & 3.00  & 3.00  & 2.98  & 3.00  \\
 & ResNet-32  & 86.94 & 86.82 & 2.34 & 3.29  & 3.26  & 3.24  & 3.22  & 3.26  & 3.24  \\
 & ResNet-56  & 88.41 & 87.93 & 2.62 & 3.66  & 3.62  & 3.65  & 3.62  & 3.63  & 3.65  \\
 & ResNet-110  & 88.33 & 87.51 & 2.67 & 3.62  & 3.62  & 3.72  & 3.70  & 3.65  & 3.72  \\
    \midrule
    \multirow{4}{*}{CIFAR-100-LT} 
 & ResNet-20  & 56.81 & 56.71 & 1.46 & 1.48  & 1.48  & 1.48  & 1.48  & 1.47  & 1.82  \\
 & ResNet-32  & 58.43 & 58.59 & 1.59 & 1.59  & 1.59  & 1.59  & 1.60  & 1.59  & 1.60  \\
 & ResNet-56  & 60.42 & 59.96 & 1.62 & 1.63  & 1.63  & 1.63  & 1.63  & 1.63  & 1.63  \\
 & ResNet-110  & 62.88 & 61.99 & 1.63 & 1.64  & 1.64  & 1.64  & 1.64  & 1.63  & 1.64  \\ \bottomrule
\end{tabular}
}
\caption{Comparison of importance weight estimators.The source is obtained with an \textbf{$\operatorname{IF}=2$}. We measure $L_1$ CE.}
\label{table:cifarlt-imagenetlt-all-estimators-0.5-p=1}
\end{table*}

\subsubsection{Experiments on real world datasets}

In Table~\ref{table:iwildcam-estimators} we report accuracy, ground truth CE (using labels) and estimated CE using different importance weight estimators on WILDS datasets: iWildCam and Amazon. The subscripts $s$ and $t$ denote the source and target distributions, respectively. Across all settings, we observe that the CE estimator using RLLS consistently yields values close to the ground truth ($\operatorname{CE}_t$). Conversely, all other weight estimation methods provide poor estimates in at least one case.

\begin{table*}[t]
\centering
\resizebox{1.0\textwidth}{!}{
\begin{tabular}{ccccccc>{\columncolor{method}}c>{\columncolor{method}}c>{\columncolor{method}}c>{\columncolor{method}}c} \toprule
   {Dataset} & {Model} & {Accuracy$_s$} &  {Accuracy$_t$} & {$\operatorname{CE}_s$} & {$\operatorname{CE}_t$} & {$\widehat{\operatorname{CE}}_t(\hat{\bo}^*)$ } & \makecell{\textbf{$\widehat{\operatorname{CE}}_t (\hat{\bo})$} \\ $\text{RLLS}$} & 
   \makecell{\textbf{$\widehat{\operatorname{CE}}_t (\hat{\bo})$} \\ $\text{ELSA}$} & 
   \makecell{\textbf{$\widehat{\operatorname{CE}}_t (\hat{\bo})$} \\ $\text{EM-BCTS}$} & 
   \makecell{\textbf{$\widehat{\operatorname{CE}}_t (\hat{\bo})$} \\ $\text{BBSL}$} \\ 
   \midrule
    \multirow{4}{*}{Amazon}
 & RoBERTa & 74.11 & 56.26 & 0.92 & 2.27 & 2.19 & 2.02 & 2.03 & 2.05 & 2.02 \\
 & D-RoBERTa & 73.09 & 56.99 & 1.71 & 3.50 & 3.41 & 3.36 & 3.41 & 2.96 & 3.36 \\
 & BERT & 73.02 & 53.46 & 0.29 & 1.25 & 1.30 & 1.06 & 1.23 & 4.03 & 1.06 \\
 & D-BERT & 71.17 & 55.03 & 2.37 & 4.37 & 4.65 & 4.41 & 4.51 & 3.83 & 4.41 \\
    \midrule
    \multirow{4}{*}{iWildCam} 
 & ResNet-50 & 85.21 & 67.80 & 0.66 & 0.99 & 1.09 & 1.13 & 1.93 & 3.39 & -- \\
 & ViT-Large & 83.23 & 63.21 & 0.75 & 1.08 & 1.26 & 1.38 & 16.81 & 3.30 & -- \\
 & ViT-Large (384) & 85.74 & 65.65 & 0.77 & 1.14 & 1.23 & 1.27 & 2.17 & 3.60 & -- \\
 & Swin-Large & 86.34 & 66.96 & 0.73 & 1.07 & 1.09 & 1.17 & 1.87 & 3.48 & -- \\ \bottomrule
\end{tabular}
}
\caption{Comparison of different importance weight estimators on Amazon and iWildCam. The target data follows a uniform distribution over the classes, which we obtained by resampling the respective test sets. We measure $L_2$ CE. We do not report numbers using BBSL on iWildCam, due to the source distribution containing 0-frequency classes, which yields the confusion matrix non-invertable.}
\label{table:iwildcam-estimators}
\end{table*}

\end{document}